\documentclass[sigconf, nonacm]{acmart}

\usepackage{packages}
\usepackage{macros}

\newcommand\vldbdoi{XX.XX/XXX.XX}
\newcommand\vldbpages{XXX-XXX}
\newcommand\vldbvolume{17}
\newcommand\vldbissue{1}
\newcommand\vldbyear{2024}
\newcommand\vldbauthors{\authors}
\newcommand\vldbtitle{\shorttitle} 
\newcommand\vldbavailabilityurl{URL_TO_YOUR_ARTIFACTS}
\newcommand\vldbpagestyle{plain}

\begin{document}

% Revision response
% \input{body/cover_letter.tex}

\title{Accelerating String-key Learned Index Structures \\ via Memoization-based Incremental Training}

%%
%% The "author" command and its associated commands are used to define the authors and their affiliations.

% TODO: Fill author information
%\affiliation{%
%  \institution{Institute for Clarity in Documentation}
%  \streetaddress{P.O. Box 1212}
%  \state{Ireland}
\author{Minsu Kim}
\affiliation{%
    \institution{KAIST}
}
\email{mskim@casys.kaist.ac.kr}

\author{Jinwoo Hwang}
\affiliation{%
    \institution{KAIST}
}
\email{jwhwang@casys.kaist.ac.kr}

\author{Guseul Heo}
\affiliation{%
    \institution{KAIST}
}
\email{gsheo@casys.kaist.ac.kr}

\author{Seiyeon Cho}
\affiliation{%
    \institution{KAIST}
}
\email{sycho@casys.kaist.ac.kr}

\author{Divya Mahajan}
\affiliation{%
    \institution{Georgia Tech}
}
\email{divya.mahajan@gatech.edu}

\author{Jongse Park}
\affiliation{%
    \institution{KAIST}
}
\email{jspark@casys.kaist.ac.kr}

\begin{abstract}

Learned indexes use machine learning models to learn the mappings between keys and their corresponding positions in key-value indexes.
These indexes use the mapping information as training data.
Learned indexes require frequent retrainings of their models to incorporate the changes introduced by update queries.
To efficiently retrain the models, existing learned index systems often harness a linear algebraic QR factorization technique that performs matrix decomposition.
This factorization approach processes all key-position pairs during each retraining, resulting in compute operations that grow linearly with the total number of keys and their lengths.
Consequently, the retrainings create a severe performance bottleneck, especially for variable-length string keys, while the retrainings are crucial for maintaining high prediction accuracy and in turn, ensuring low query service latency.
To address this performance problem, we develop an algorithm-hardware co-designed string-key learned index system, dubbed \lia.
In designing \lia, we leverage a unique algorithmic property of the matrix decomposition-based training method.
Exploiting the property, we develop a memoization-based incremental training scheme, which only requires computation over updated keys, while decomposition results of non-updated keys from previous computations can be reused.
We further enhance \lia to offload a portion of this training process to an FPGA accelerator to not only relieve CPU resources for serving index queries (i.e., inference), but also accelerate the training itself. 
Our evaluation shows that compared to ALEX, LIPP, and SIndex, a state-of-the-art  learned index systems, \sia-accelerated learned indexes  offer 2.6$\times$ and 3.4$\times$ higher throughput on the two real-world benchmark suites, YCSB and Twitter cache trace, respectively.

\end{abstract}

\maketitle

\setcounter{page}{1}

\pagestyle{\vldbpagestyle}
\begingroup\small\noindent\raggedright\textbf{PVLDB Reference Format:}\\
\vldbauthors. \vldbtitle. PVLDB, \vldbvolume(\vldbissue): \vldbpages, \vldbyear.\\
\href{https://doi.org/\vldbdoi}{doi:\vldbdoi}
\endgroup
\begingroup
\renewcommand\thefootnote{}\footnote{\noindent
This work is licensed under the Creative Commons BY-NC-ND 4.0 International License. Visit \url{https://creativecommons.org/licenses/by-nc-nd/4.0/} to view a copy of this license. For any use beyond those covered by this license, obtain permission by emailing \href{mailto:info@vldb.org}{info@vldb.org}. Copyright is held by the owner/author(s). Publication rights licensed to the VLDB Endowment. \\
\raggedright Proceedings of the VLDB Endowment, Vol. \vldbvolume, No. \vldbissue\ %
ISSN 2150-8097. \\
\href{https://doi.org/\vldbdoi}{doi:\vldbdoi} \\
}\addtocounter{footnote}{-1}\endgroup

\ifdefempty{\vldbavailabilityurl}{}{
\vspace{.3cm}
\begingroup\small\noindent\raggedright\textbf{PVLDB Artifact Availability:}\\
The source code, data, and/or other artifacts have been made available at \url{https://github.com/sia-index/sia}.
\endgroup
}

\section{Introduction}
\label{sec:introduction}

% \highlight{
% Papers about learned index in VLDB.~\cite{FINEdex-21, NFL-21, li-benchmark, PLIN-22, FILM-22, lipp, APEX-21, tsunami, learned-index-benefits, LIDER-22, DILI-23, learned-index-comprehensive, LMSFC-23}
% %
% Papers about learned index in SIGMOD.~\cite{alex, learnedindex, poisoning-learned-index, lisa, cdfshop, WISK-23, learned-index-disk-dbms}
% %
% Papers about string key value store.~\cite{forestdb, sindex, burst-tries, str-key-icde20, cuckoo-trie, wormhole, parallax, PINK-20, recipe-19, treeline-22, cassandra-trie, rosetta-20}
% }

Machine learning for system infrastructure is growing particularly in areas where data-driven decisions can make meaningful strides~\cite{learning-memory-access, learning-compiler, learning-tensor}.
Efficient data access is one such avenue, where learning indexes have proven to be effective and practical~\cite{xindex, sindex, FINEdex-21, NFL-21, li-benchmark, PLIN-22, FILM-22, lipp, APEX-21, tsunami, learned-index-benefits, LIDER-22, DILI-23, learned-index-comprehensive, LMSFC-23, alex, learnedindex, poisoning-learned-index, lisa, cdfshop, WISK-23, learned-index-disk-dbms, multidimensional-index, lidusa, sprig, plex, learned-index-google-deployment}. 
%
%Traditionally, general-purpose techniques like B-Trees and hashmaps have been widely used by data management systems to index data.
%
%However, real-world systems often exhibit unique data patterns, which allow for the replacement of these general-purpose techniques with machine learning methods.
%
The pioneering work~\cite{learnedindex} proposed in this space uses a collection of machine learning models to create a read-only ordered index for integer keys.
Due to its popularity and applicability, numerous follow-up research projects have extended the initial work to support read-write (updatable) indexes~\cite{xindex, sindex, alex, FINEdex-21, NFL-21, lipp, DILI-23, learned-index-disk-dbms, dytis-23}, string keys~\cite{sindex, Spector2021BoundingTL, XStore-20}, multi-dimensional indexes~\cite{multidimensional-index, tsunami, LIDER-22, LMSFC-23}, spatial indexes~\cite{lisa, lidusa, sprig, WISK-23}, and other variants~\cite{plex, PLIN-22, FILM-22, APEX-21}.
This paper focuses on identifying performance challenges of \emph{updatable string-key} learned indexes and addressing the challenges through an algorithm-hardware co-designed solution.

\begin{figure}[t]
        \centering
        \includegraphics[width=0.75\linewidth]{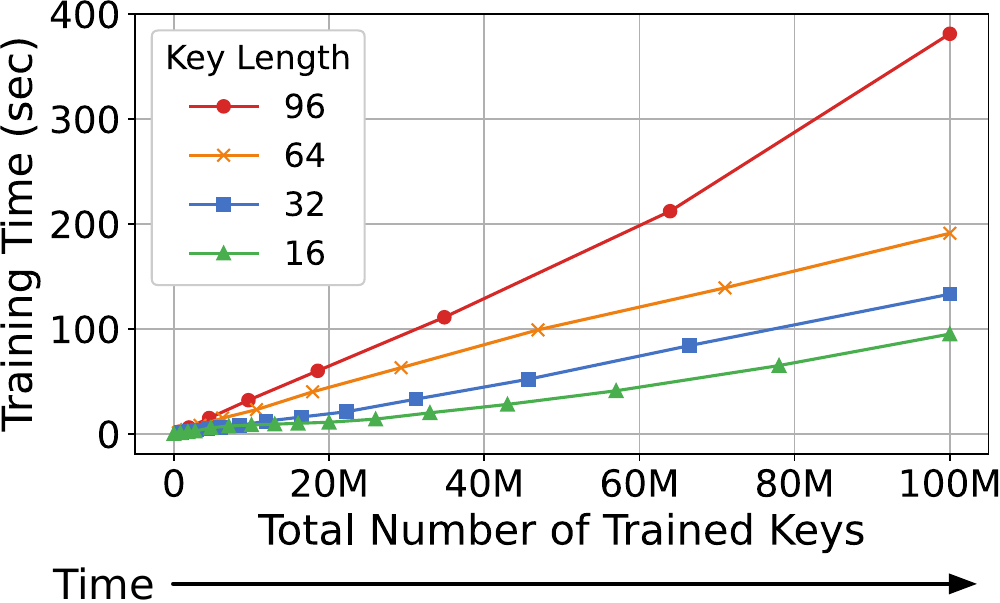}
        \vspace{-2ex}
        \caption{Increasing retraining time as the size of a learned index system grows, resulting from a stream of update queries. Markers on the same line represent sequential retraining runs, where markers positioned to the left precede those on the right. We use various key lengths - 16, 32, 64, and 96.}
        \vspace{-5ex}
        \label{fig:scalability}
\end{figure}

% In the learned index approach, machine learning models, such as linear regression, are employed to identify the correlation between keys and the positions of the records they reference.
%
Regardless of the data types of keys, an algorithmic commonality among most existing learned indexes is that the indexes are constructed as a hierarchical structure where each node is a linear model~\cite{learnedindex, xindex, sindex, alex, lipp, bourbon, colin, FINEdex-21, NFL-21, PLIN-22, APEX-21, DILI-23, tsunami}.
These linear models are designed to collaboratively \emph{learn} the mappings between keys and their corresponding positions, using this information as training data.
The training process is inherently repetitive since the key-position mappings constantly change due to the update queries (e.g., insert or delete), which necessitates \emph{retrainings} to incorporate the changes into the models.
In learned indexes, training of linear models is essentially solving the following linear equation, $X{\beta}{=}Y$ where $X$ is a key matrix, $\beta$ is a learnable parameter vector, and $Y$ is the corresponding position vector.
When learned indexes only support integer keys, the training process is computationally trivial since $X$ is a vector of integer key values (i.e., $n{\times}1$ matrix). 
However, when the keys are variable-length strings, $X$ becomes a $n{\times}k$-size matrix where $k$ is the key length, which makes solving the equation a computationally non-trivial task.
To algorithmically reduce the compute load of this training, existing string-key learned indexes~\cite{sindex, Spector2021BoundingTL, XStore-20, learnedindex} employ a matrix factorization strategy known as \emph{QR decomposition}, which enables training to be free from the burdens of matrix inversion.
Despite the algorithmic optimization, we observe that in the existing systems, the repetitive retrainings incur a severe performance bottleneck, since (1) the complexity of QR decomposition, although lower than matrix inversion, remains high, and (2) retrainings and index query servicing for existing keys (i.e., inference) compete for the limited CPU resource. 
%
% During the training process, the models are intentionally overfit to the training data, which differs from traditional machine learning models, where the goal is to generalize to unseen inputs.
%
% Indexing structures only encounter keys that have been inserted, and thus, they can benefit from overfitting to the training data. 
%
% To efficiently train the models, learned indexes often leverage a \emph{matrix decomposition} based technique for each linear model training and obtain the parameters of the regression models using the inverses of the matrices.
%
%
% When a learned index system receives update queries (e.g., insert or delete), it needs a \emph{retraining} to incorporate the changes into the models, while concurrently serving read queries for existing keys (i.e., inference).
%
% We observe that in existing learned index systems, these repetitive retrainings cause a performance bottleneck due to the substantial computational cost.
%
Figure~\ref{fig:scalability} shows that retraining time progressively grows as the number of keys and key lengths increase, on a state-of-the-art string-key learned index, SIndex~\cite{sindex}.
Each point in the graph represents a retraining run.
Increased retraining times negatively impact the inference throughput, as they result in an outdated index.
This, in turn, lowers the index prediction accuracy and necessitates a costly linear search to locate the correct position.
Thus, retraining is crucial for reducing service latency as well as improving index throughput.
To address the aforementioned bottlenecks, we introduce \textbf{SIA}: \textbf{S}tring-key Learned \textbf{I}ndex \textbf{A}cceleration.
\sia enables efficient and scalable indexing by reducing the compute load of the retraining process through an algorithmic technique and judiciously offloads a portion of the training computation onto an FPGA accelerator.
The challenge is that current learned indexes need to perform costly matrix decomposition using the \emph{entire} key-position mappings as input to maintain model accuracy, which is pivotal for achieving high index performance.
To tackle this challenge, \sia utilizes a modified parallel decomposition technique that allows for piecewise computation of matrix decomposition.
In designing \sia, we leverage the insight that these retrainings occur on progressively updated indexes, thus offering an opportunity to reuse computations from prior results via memoization. 
It is important to note that training using the memoized decomposition results produces mathematically identical outcomes to those obtained if the models were fully retrained from scratch using the complete set of keys.
Building on the memoization-based decomposition, we develop a learned index training algorithm that incrementally retrains the models by leveraging the results of prior matrix decomposition.
This enhanced algorithm reduces the computational complexity and retraining time, which in turn frees up CPU resources for servicing queries.
However, our empirical analyses suggest that the algorithmic optimization, while helpful, offers a limited benefit since the retrainings still compete over the limited CPU resource.
To further reduce the retraining time, we enable the retrainings to be accelerated using an FPGA.
We choose FPGA over GPU owing to its customizability to index-specific algorithm configurations, leading to enhanced energy efficiency.
\sia combines these elements to offer a novel learned index mechanism that aims to improve system query throughput through both algorithmic and hardware innovations.
This work makes the following contributions: 

\begin{itemize}[itemsep=0pt, parsep=0pt, topsep=1pt, partopsep=1pt]
    \item 
    %Identifies the system bottlenecks in current updatable learned index structures for string-keys, specifically, retraining the ensemble models in the hierarchical structure. We observe that as the retraining time grows, it progressively leads to longer execution latency for both read and update queries.
    %
    Identifies the system bottlenecks in current updatable learned index structures for string-keys, specifically, retraining the ensemble models in the hierarchical structure. We observe that as the retraining time grows, it progressively leads to lower performance of learned index systems.
    \item Introduces a novel learned index system, \sia, that accelerates the retraining process through an enhanced mathematical approach to matrix decomposition, enabling incremental training. With incremental training, only updated keys are used for computation, while the computation result for old keys is reused.
    \item Further accelerates \sia's incremental training process using an FPGA-based design that reduces training time and frees up CPU resources for index query servicing.
\end{itemize}

We demonstrate the effectiveness of \sia using two real-world benchmark suites, YCSB and Twitter cache trace.
%, along with a microbenchmark for sensitivity study.
%
For YCSB, we use two datasets available to the public, Amazon review and MemeTracker datasets, as well as a synthetic dataset.
We integrate \sia into the three recent updatable string-key learned indexes, including ALEX~\cite{alex}, LIPP~\cite{lipp}, and SIndex~\cite{sindex}.
Compared to baseline learned indexes, \sia-accelerated learned indexes provide 2.6$\times$ and 3.4$\times$ higher throughput for YCSB and Twitter cache trace workloads, respectively.
From an in-depth ablation study using SIndex that breaks down the benefits of \sia, we observe that employing solely the memoization decomposition-based incremental learning algorithm offers 1.6$\times$ and 1.9$\times$ higher throughput.
However, when the FPGA-based \sia accelerator is employed, it offers 2.8$\times$ and 4.3$\times$ higher throughput than the baselines, which are respectively 1.8$\times$ and 2.3$\times$ \emph{additional} speedup, a substantial performance boost compared to the software-only counterpart.
These results suggest that taking an algorithm-hardware co-design approach, \sia enables heterogeneous CPU-FPGA architecture to operate as a platform of choice to achieve high throughput for updatable string-key learned indexes.
\section{A Primer on Learned Index}
\label{sec:background}

\begin{figure*}[t]
        \centering
        \includegraphics[width=0.95\linewidth]{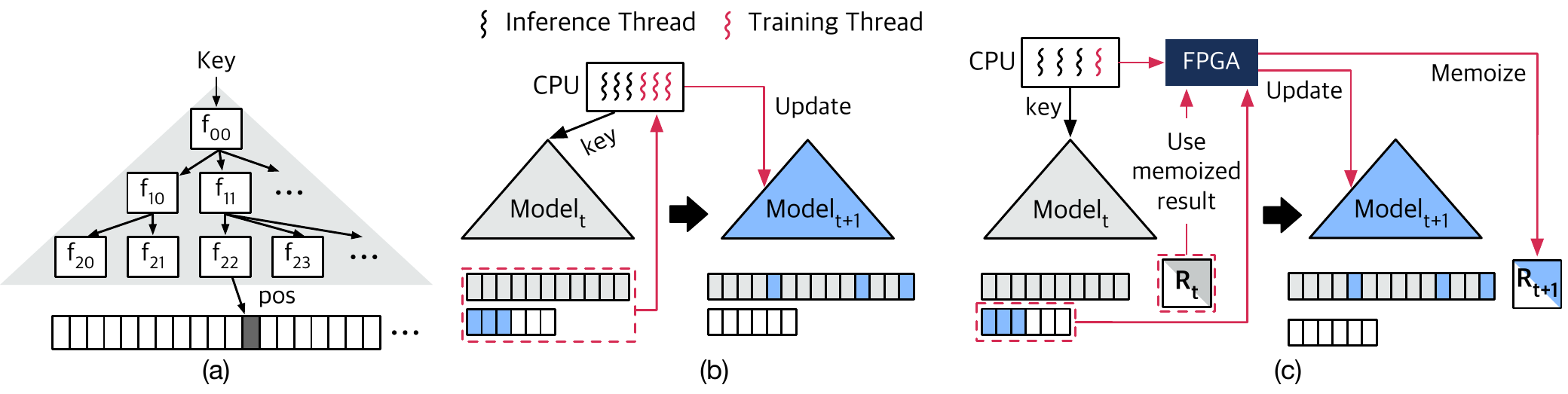}
        \vspace{-3ex}
        \caption{(a) Read-only learned index in a hierarchical structure, (b) updatable learned index, and (c) \lia: the proposed updatable string-key learned index that leverages computation reuse and hardware acceleration to improve the system throughput.}
        \vspace{-1ex}
        \label{fig:overview}
\end{figure*}

\begin{figure}[t]
        \centering
        \includegraphics[width=1.0\linewidth]{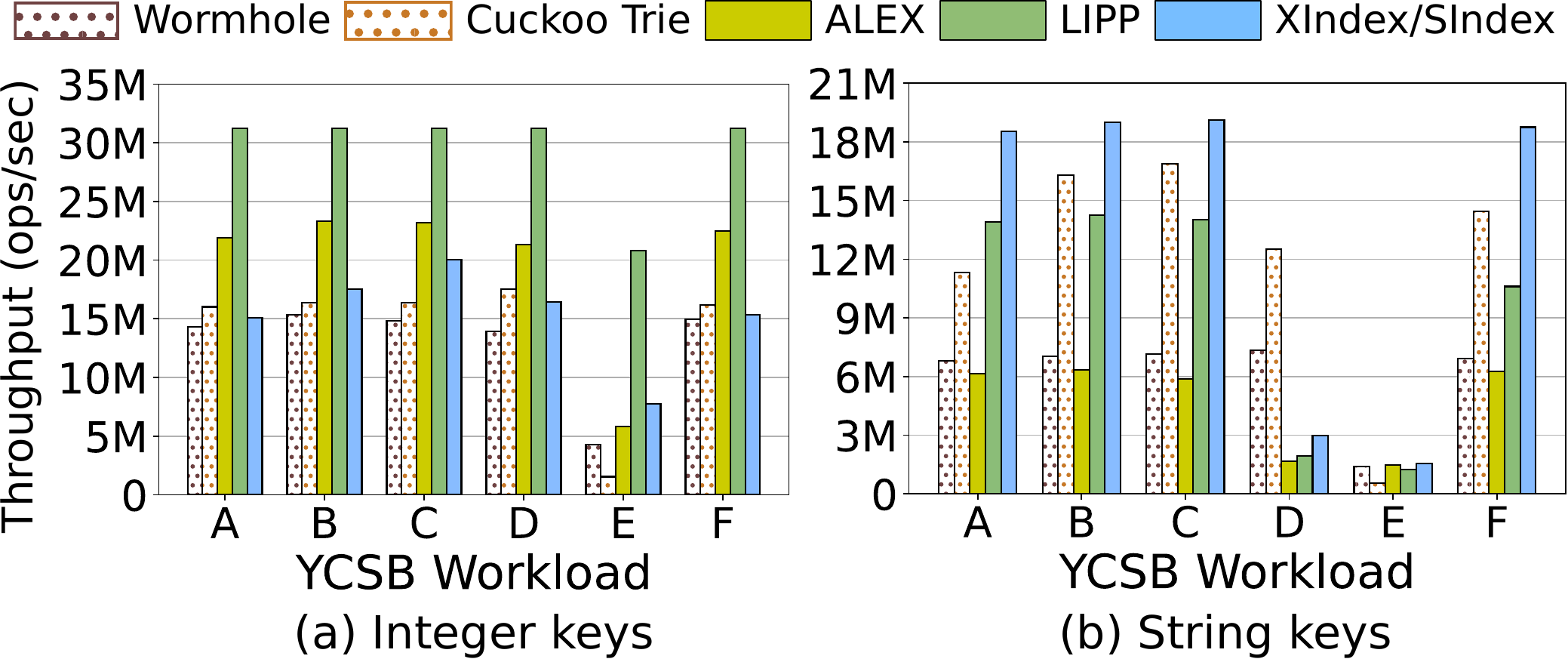}
        \vspace{-5ex}
        \caption{Throughput results of two conventional indexes and three learned indexes for YCSB workloads.}
        \vspace{-4ex}
        \label{fig:li-intstr}
\end{figure}

%Index is a common data structure used in data management applications.
%
Key-value stores are widely deployed in data management applications, where the index maps keys to their corresponding positions in a list of records.
This pairing can be denoted as a function, $f(key, \ position)$, with key as the input and position as the output.
Conventionally, hash-map and B-tree structures are commonly used to store this mapping in an array of records.
Despite its popularity, they still have shown several limitations, which prevent their ``one-size-fits-all'' deployment.
While hash-maps typically offer low average access time, they can be susceptible to hash collisions that may lead to unpredictable increases in lookup and construction time.
Additionally, hash-maps may not perform as well as other data structures for range queries.
On the other hand, B-trees and their variants do not have the same limitations as hash-maps, but their average-case performance, in terms of both latency and throughput, is generally lower than that of hash-maps.
To overcome these limitations, the community has explored the use of machine learning (ML) approaches to develop learned indexes as an alternative index structure~\cite{learnedindex, sindex, xindex, pgm-index, alex, lisa, lipp, tsunami, li-benchmark, bourbon}.

\niparagraph{Learning to index keys.}
The initial work on learned indexes~\cite{learnedindex} demonstrated that it is possible for ML models to learn the mappings between keys and their corresponding positions, using this information as training data.
Unlike traditional machine learning models, which aim to generalize to unseen inputs, learned index models are intentionally overfit to the training data, as index structures mostly encounter keys that have been inserted.
%
%During the training process, the models are intentionally overfit to the training data, as opposed to traditional machine learning models, where the goal is to generalize to unseen inputs. 
%
%Indexing structures mostly encounter keys that have been inserted, and thus, they can benefit from overfitting to the training data.
%
Despite the overfitting during training, inference-based indexing can still produce incorrect predictions due to the inherent approximation nature of machine learning.
When the queried key is not found at the predicted position, learned indexes search for the key within a bounded range around the predicted position (i.e., [$p$ + $err_{min}$, $p$ + $err_{max}$]), ensuring accurate indexing functionality~\cite{learnedindex}.
When designing a ML model for learned indexes, there are various alternatives that trade off accuracy (model size and architecture) against cost (inference latency and training time). 
Several works~\cite{learnedindex, sindex, xindex, pgm-index} have shown that a hierarchical structure of linear models effectively balances this tradeoff.
Each node in the hierarchical structure is a linear regression model that needs to be trained for a subset of the key-position mappings.
%
%Our extensive literature survey suggests that currently over 85\% of existing learned index~\cite{learnedindex, xindex, sindex, alex, lipp, bourbon, colin, FINEdex-21, NFL-21, PLIN-22, APEX-21, DILI-23, tsunami} works rely on linear regression models, as they have sufficient learning capabilities for ordered keys and offer simplicity that incurs minimal training and inference costs.
%
%\textcolor{red}{In this paper, the training time implies the training of the entire RMI model and not just a single node of linear regression model.}
%
%This hierarchical structure allows RMI to accurately learn the key-address mappings using linear regression as the learning algorithm.
%
% Further, each leaf node is equipped with two parameters that bound the errors, $err_{max}$ and $err_{min}$, which guarantee that, if the record exists, the queried key is positioned within the range [$p$ + $err_{min}$, $p$ + $err_{max}$]. 
%
These initial works focus on read-only indexes, and hence training is carried out once when building the indexes before deployment. 
The hierarchical structure for read-only indexing is depicted in Figure~\ref{fig:overview}(a).
%
%Moreover, \textsf{Learned Index} is designed for integer keys, which makes training the linear regression models a trivial job (i.e., solving $y = ax + b$ where both $x$ and $y$ are integer vectors).
%

\niparagraph{Updatable string-key learned index.}
Although restricting the scope to \emph{read-only} indexes was an effective setting to demonstrate initial applicability of the ``learning'' approach, practical data management necessitates support for ``update'' queries (e.g., \texttt{insert} and \texttt{delete}).
Follow-up works overcome this limitation and devise ``updatable'' learned indexes~\cite{alex, lipp, xindex, sindex}.
%
%PGM~\cite{pgm-index} leverages log-structured merge (LSM) tree to incrementally append newly inserted keys to differently-sized subsets and builds learned index structures on top of these subsets.
%
ALEX~\cite{alex} expands nodes with deliberately-reserved empty spaces for unseen future keys, which hold the newly inserted keys until the updated keys are retrained.
LIPP~\cite{lipp} ensures precise model prediction results and removes costly local search usually used in other learned indexes.
XIndex~\cite{xindex} is another variant that maintains reserved spaces for future keys, while unlike ALEX, the new keys are stored in separate temporary buffers.
SIndex~\cite{sindex} is one of the initial efforts to support variable-length string keys in learned indexes.
As string keys are an important datatype used in diverse applications such as web servers, sequence analysis, and genomics, modern key-value stores often have strong support for this datatype~\cite{forestdb, sindex, burst-tries, str-key-icde20, cuckoo-trie, wormhole, parallax, PINK-20, recipe-19, treeline-22, cassandra-trie, rosetta-20}.
Despite its importance, its performance implication on updatable string-key learned index systems remains under-examined in existing literature, which is the primary focus of this work. 
%

%
%Therefore, the variable-length keys are a critical feature for indexes to be used in practical settings.  
%
%There are also other works~\cite{lipp, OTHERS} that propose algorithmic and system techniques to improve the throughput and latency of learned index. 
%
% LIPP~\cite{lipp} identifies the limited model prediction accuracies of these prior works as the main performance degradation factor and tackles this challenge by restructuring the learned index management scheme such that the predictions are always precise. 

\niparagraph{Intertwinement of retraining and inference.}
Unlike traditional machine learning, training and inference phases in these updatable learned indexes are not clearly demarcated. 
Instead, learned indexes require iterative \emph{retrainings}, because the training data is constantly changing due to update queries.
%
%Instead, learned indexes require constant and iterative \emph{retrainings}, because the training data that the models need to overfit is constantly changing due to update queries.
%
Concurrently, the index systems must serve index queries by performing \emph{inference}. 
This convergence of training and inference can influence each other's performance, potentially resulting in a marked degradation of overall efficiency. 
Figure~\ref{fig:overview}(b) delineates the common execution flow where certain threads are dedicated to query servicing and certain to retraining.
\niparagraph{Effectiveness of learned indexes.}
To better understand the effectiveness of learned indexes, we conduct preliminary experiments comparing the throughput of learned index structures with two \emph{non}-learned indexes, Wormhole~\cite{wormhole} and Cuckoo Trie~\cite{cuckoo-trie}.
We use the Yahoo! Cloud Serving Benchmark (YCSB)~\cite{ycsb}, a key-value store benchmark suite with six different workloads
%, each associated with a different query and key distribution, intended to simulate real-world cloud-based data serving systems for various applications such as social media, e-commerce, and web search 
(see Section~\ref{sec:method} for details).
Figure~\ref{fig:li-intstr} shows that learned indexes generally offer comparable or higher performance than the two baseline indexes for both integer and string key cases.
%
% For integer keys, most workloads exhibit similar trends except the workload \textsf{E}, which has a mix of short-range scans and inserts, and causes Cuckoo Trie to perform poorly due to its reliance on hash-maps.
%
However, the notable observation is that when keys are string, learned indexes perform much worse than the baseline indexes for workload \textsf{D} and \textsf{E}.
Workload \textsf{D} and \textsf{E} contain \texttt{insert} queries, which necessitate the constant retrainings for index updates.
While these retrainings impose marginal performance overhead when keys are integers, retrainings for string keys become severe performance bottleneck, 
cancelling the performance benefits of learned indexes, as will be deeply analyzed in Section~\ref{sec:motivation}.
This is the very challenge we aim to tackle in this work through \sia.

\niparagraph{Our approach to tackle the challenge.}
\lia sets out to tackle the challenges posed by current updatable string-key learned indexes, with the following objectives: (1) \lia aims to reduce the cost of training linear models without any mathematical implication on model quality, and (2) it aims to enhance the system with an FPGA accelerator that can execute the compute-intensive portion of training, thus relieving CPU resources for inference.
Figure~\ref{fig:overview}(c) depicts \lia's system architecture, which is built upon existing learned index systems.
\section{Analyses of Learned Indexes}
\label{sec:motivation}

We conduct in-depth performance characterizations through a set of preliminary experiments and obtain three main insights from the results.
These insights form the key driving forces behind \lia.
For these analyses, we use a SIndex system running on a 16-core server, the details of which are provided in Section~\ref{sec:method}.
We use a workload with uniformly distributed keys, generating \texttt{read} and \texttt{insert} queries based on a predetermined ratio (e.g., 70\% \texttt{read} and 30\% \texttt{insert} queries).
%
%We use a microbenchmark that generates \texttt{read} and \texttt{insert} queries based on a predetermined ratio (e.g., 70\% \texttt{read} and 30\% \texttt{insert} queries).
%
\texttt{Insert} queries raise the retraining complexity by adding more keys to the index.
We initialize the index with 1M keys.

\subsection{Retraining-Time Scalability Analysis}
\label{sec:scalability-analysis}

\if 0
\begin{figure}[t]
        \centering
        \includegraphics[width=0.8\linewidth]{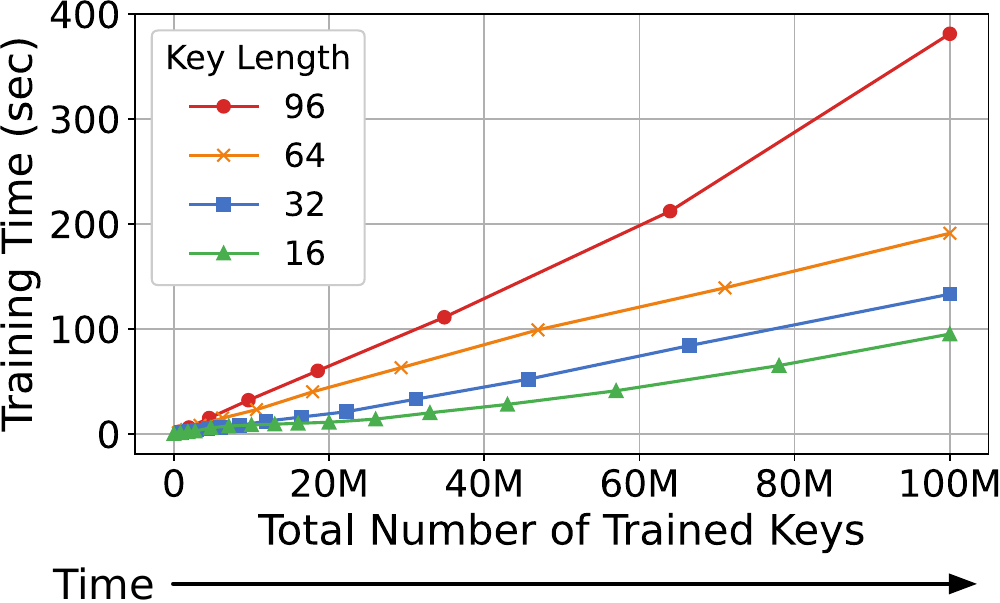}
        \caption{Training time as the number of keys increases. Each marker represents a training run. The experiments are performed for different key lengths that range from 16 to 32, 64, and 96.}
        \label{fig:scalability}
\end{figure}
\fi 

%
Existing updatable string-key learned indexes suffer from a limitation in that they aggregate all keys into a single dataset, making computations more demanding as the number of keys increases.
To examine the scalability aspect of learned indexes, we measure the retraining time as we gradually increase the total number of keys from 1M to approximately 100M. 
Figure~\ref{fig:scalability} shows the results with each marker representing a retraining invocation.
The experiment shows that for total numbers of keys reaching 100M, the retraining time for learned indexes becomes prohibitively long. 
Retraining time for the shortest key length of 16 increases up to 100 seconds, while it exceeds 5 minutes for the case of key length 96.
These extended retraining times for indexing are infeasible as they result in the index being significantly outdated.
The results also show that progressively prolonged retraining time ends up leading to longer intervals between retraining invocations.
This delay occurs because the growing retraining time increases the number of keys waiting for the next round of retraining, resulting in a lower frequency of model updates.

\finding{This analysis shows that the existing updatable string-key learned index systems face scalability issues. Thus, there is a need for a solution that minimizes the retraining time for linear models, especially when dealing with large index sizes and long key lengths.}

\begin{figure}[t]
        \centering
        \includegraphics[width=0.75\linewidth]{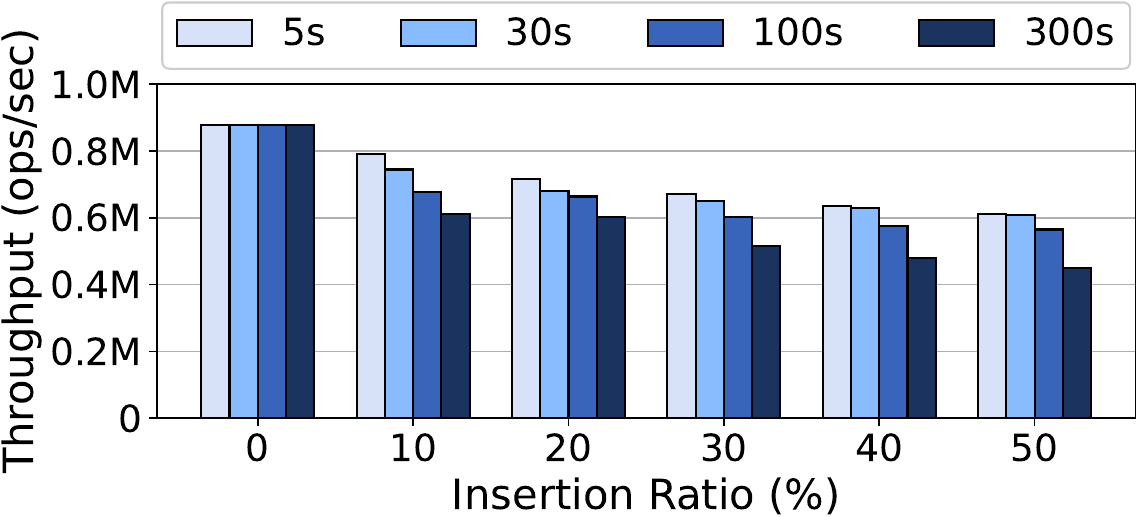}
        \vspace{-2ex}
        \caption{Throughput as training time varies from 5 to 300 seconds. Training does not utilize any CPU cycles. The insertion ratio sweeps from 0\% (read-only) to 50\%.}
        \vspace{-3ex}
        \label{fig:implication-of-slow-training}
\end{figure}

\subsection{Impact of Slow Retraining on Throughput}
\label{sec:slow-training}
Given the aforementioned insight, a subsequent research question could be, ``Why is retraining vital for the overall efficacy of the learned index system?''
The response to this inquiry is that training influences throughput in two significant ways.
(1) First, slow retraining causes the models to become outdated, resulting in reduced index prediction accuracy and requiring a costly linear search to locate the correct position. 
This, in turn, leads to prolonged index search latencies for more \texttt{read} queries, negatively impacting the overall system throughput.
(2) Second, as retraining and inference run simultaneously on the same system and compete for CPU resources, the inference throughput is adversely affected.  
We discuss the first implication in this section and leave the discussion for the second effect to Section~\ref{sec:cpu-contention}.
To demonstrate the impact of slow training over throughput, we develop a ``fictitious'' system that can retrain linear models within a predetermined training time \emph{without} using any CPU resources for training.
This method allows us to isolate the impact of slow retraining separate from the implications of CPU resource contention.
Figure~\ref{fig:implication-of-slow-training} depicts the throughput of this fictitious system as the retraining duration shifts between 5s and 300s.
%
% In this context, throughput refers to the total number of queries (both \texttt{read} and \texttt{insert}) processed per second. 
%
The results show a consistent decline in throughput as the retraining time lengthens, since learned indexes must use outdated models during the retraining period, which would increase the frequency and degree of linear search to locate the correct position.
Additionally, we observe that as the insertion ratio rises, the system sees a decline in throughput. 
This is because a greater number of inserted keys await in the buffer before integration into the learned index, which again requires more overhead on linear search at the non-trained key buffers.
While the reported throughput averages over time, in a practical scenario, throughput would gradually drop as runtime progresses, because, unlike our hypothetical system, a real system would face an ever-increasing retraining time.

\finding{Our study suggests that a long retraining period hurts the end-to-end system throughput of updatable string-key learned index systems. Therefore, fast retraining of linear models is imperative.}

\subsection{Implication of CPU Resource Allocation}
\label{sec:cpu-contention}
\begin{figure}[t]
        \centering
        \includegraphics[width=0.9\linewidth]{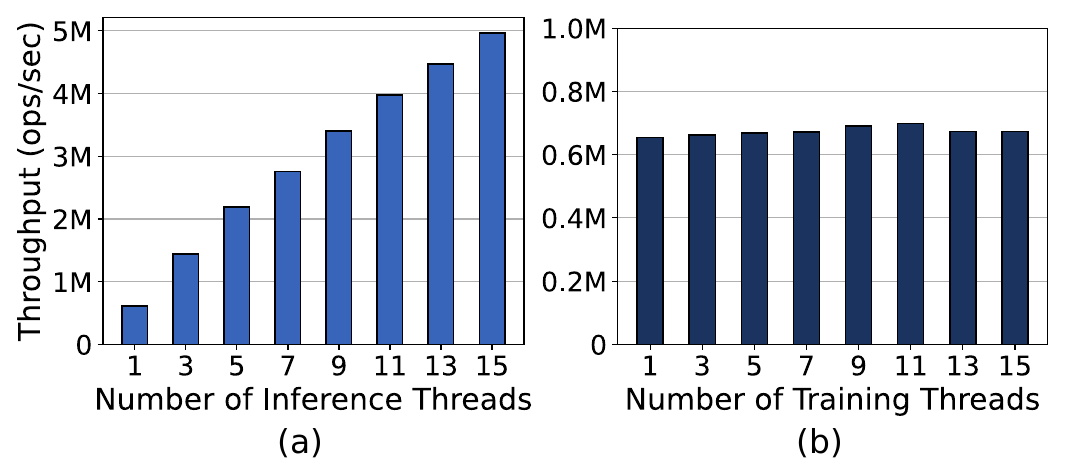}
        \vspace{-3ex}
        \caption{Throughput with varying threads for (a) inference and 1 for training, (b) training and 1 for inference.}
        \vspace{-2ex}
        \label{fig:inference-thread-scale}
\end{figure}

A straightforward solution to reduce training time would be to allocate more CPU resources. 
To better understand the correlation between throughput and CPU resources, we perform an experiment that measures the system throughput as we vary the number of threads allocated for inference (index serving) and training, while maintaining the number of threads assigned to the other task at 1.
This approach allows us to determine the performance benefits that inference and training could achieve with additional CPU resources, respectively.
As our system has 16 cores, we vary the number of cores allocated to either inference or training threads from 1 to 15, maintaining the insertion ratio at 50\% and the key length at 32. 
Figure~\ref{fig:inference-thread-scale}(a) and Figure~\ref{fig:inference-thread-scale}(b) show throughput trends. 
When the number of threads for training is 1, the additional CPU threads allocated for inference result in sub-linear yet substantial performance scaling. 
This is because inference is read-only and multiple inferences can be executed independently and in parallel across threads.
However, when the inference process is restricted to a single thread while training utilizes an increasing number of cores, the additional resources only yield marginal benefits.
The limited effectiveness of CPU for training can be attributed to the limited parallelism in the matrix decomposition algorithm used for linear regression training, as explained in further detail in Section~\ref{sec:linear-regression-training}.
%
% The ``Ideal'' bar shows again the fictitious scenario where training does not require any CPU usage and the training time is zero. 
% %
% It represents the maximum potential throughput that could be achieved if the training speed is infinitely accelerated. 
% %
% This paper seeks to achieve high throughput by adopting an algorithm-hardware co-design approach that strives to approach close to the maximum achievable throughput.
%

\finding{We note that inference gains more from extra CPU resources compared to training. As a result, we propose a heterogeneous system that allocates CPU resources primarily for inference, while employing an FPGA accelerator for the training process.}
\vspace{-2ex}

\section{\liatitle Design Principles}
\label{sec:overivew}

Building upon the insights, we propose a hardware-accelerated updatable string-key learned index system, dubbed \lia.
First, \lia proposes a novel incremental index learning algorithm, which reduces the computing complexity and execution time of each retraining process. 
\lia then dedicates most of the CPU resource for inference serving by offloading the training to an FPGA accelerator, thus collaboratively achieving high throughput. 
This section outlines the design principles of each \lia component.
\niparagraph{\emph{Algorithm} design principle: Performing only necessary computations for learned indexes.}
The fundamental challenge addressed in this work is the lack of scalability in learned index training since the compute operations for training \emph{compounds} as the number of keys grows. 
In learned index systems, every retraining run necessitates the processing of the entire dataset. 
The current state-of-the-art approach involves performing matrix decomposition, matrix inverse, dot product, and transpose operations over the \emph{entire} dataset to determine the parameters of the linear models. 
To reduce the computing complexity of the training, we devise an incremental index learning algorithm that memoizes the results of previous retraining computations and reuses them in combination with the new results obtained from the augmented training data.
With this algorithm, the computational load is not determined by the total number of keys, but rather by the number of updated keys.

\niparagraph{\emph{Hardware} design principle: Designing the accelerator specifically for training to ensure high energy efficiency as index systems are often dedicated for the exclusive purpose and are consistently operational.}
Updatable learned indexes must ceaselessly perform training to keep up with the changes made by update queries, which makes achieving high energy efficiency a primary concern in designing the systems. 
Although employing a GPU is seemingly a straightforward approach to attaining high throughput, the advantage is offset by substantial energy consumption. 
Thus, in this work, we choose FPGA as our platform.
FPGA not only allows us to customize accelerators for diverse algorithm/system constraints and thus achieve high energy efficiency, but also it is already available in the form of off-the-shelf cards, which facilitates integration with the existing systems~\cite{fpga-nic, fpga-cpu}. 
To effectively utilize FPGAs for changing training configurations and index model sizes, we develop a hand-optimized design specifically for the proposed memoization-based incremental training algorithm. 

\niparagraph{\emph{Software} design principle: Enabling plug-and-play based runtime software for generality and non-invasiveness.}
While hardware acceleration can offer significant performance gains, the proposed technique needs to be integrated seamlessly with existing learned index systems.
Thus, \lia cannot be specific to a certain updatable learned index.
\lia's system software is built by determining the commonalities of existing updatable learned indexes and integrating the FPGA-based accelerator with minimal modifications to the existing software stack.
To accomplish this objective, we utilize the fact that although various learned indexes may have different model structures and index management mechanisms, they all rely on linear regression models as the fundamental kernel, which can be readily separable from the other components of the index system.
Given this insight, we encapsulate the accelerator and its driver as a linear model training library, which is customized for the case where the training data incrementally grows or shrinks.
%
%As such, our acceleration solution is not specific and is in general applicable to any updatable learned indexes.

%\bluetext{The rest of this paper presents how we follow these principles to design the three components of \lia: (1) the incremental index learning algorithm, (2) an FPGA-based accelerator customized for the proposed learning algorithm, and (3) a plug-and-play library centered on linear model-based training for seamless integration.}

\section{Incremental Index Learning}
\label{sec:algorithm}

\begin{figure}
        \centering
        \includegraphics[width=1\linewidth]{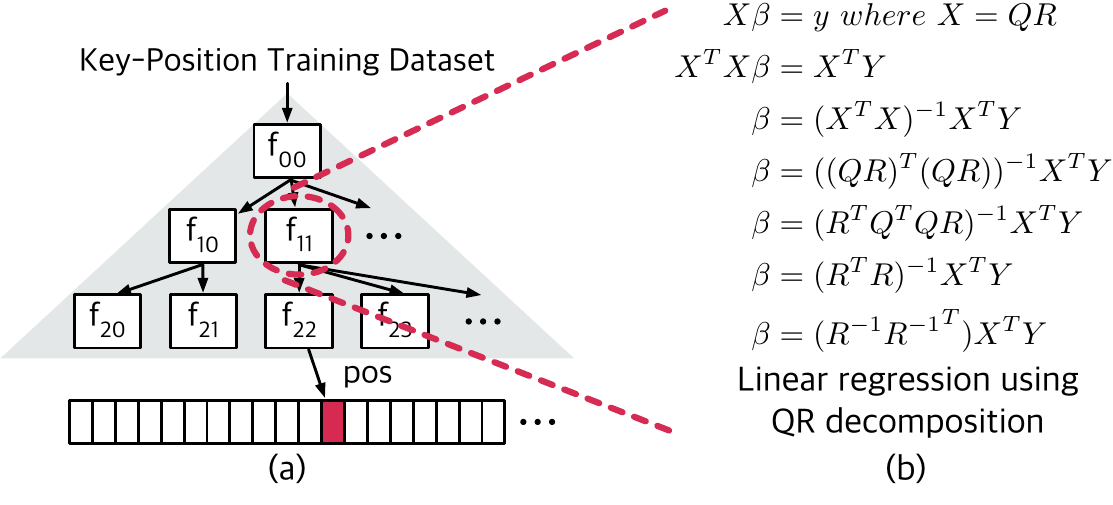}
        \vspace{-5ex}
        \caption{(a) Recursive Model Index, and (b) Linear regression training operations required per model.}
        \vspace{-3ex}
        \label{fig:index_training_process}
\end{figure}

Reducing the training workload of the updatable learned index structures is a key challenge tackled by this work.
In this section, we first provide the background on training hierarchically structured learned indexes, which requires linear regression training using matrix decomposition.
We then introduce \lia's novel index learning algorithm, which effectively reduces the computational load of the training process via reuse, without any changes to model quality.

\subsection{Hierarchical Model Index Training}
\label{sec:rmi}
Most learned indexes~\cite{learnedindex, xindex, sindex, alex, lipp, FINEdex-21, NFL-21, APEX-21, radixspline, LIDER-22, DILI-23, FILM-22, pgm-index} share a unique commonality by employing a hierarchical model index structure, as illustrated in Figure~\ref{fig:index_training_process}(a).
In the hierarchical structure, the internal and leaf nodes have different roles: \emph{learned models at internal nodes} predict which node to traverse among the children and \emph{learned models at leaf nodes} predict the positions for the queried keys. 
This structure splits the entire key range into a series of small and possibly overlapping ranges, where each range is assigned to a leaf node and learned with the associated model.
Note that due to update queries, the index structure can potentially expand or shrink, as the total number of keys handled by the system increases and decreases. 
The hierarchical index is trained in two main ways: (1) cold training from scratch, which is for new nodes created due to the index structure changes, and (2) updating pre-existing models within the existing nodes due to key additions or deletions without any alterations to the hierarchical index structure.
Cold training is infrequent, typically triggered only when the prediction accuracy falls below a set threshold.
Mostly, keys are updated without the need to add or remove any nodes.
Hence, \lia focuses on optimizing the latter, reserving conventional training techniques for the former.

\subsection{Linear Regression Training}
\label{sec:linear-regression-training}

Linear regression (LR) models the relationship between variables by fitting a linear equation to training data.
Formally, given an input $X = ((x_{11}, ..\ , x_{1p}), ..\ , (x_{n1}, ..\ , x_{np}))$ and output $Y = (y_1, ..\ , y_n)$, a LR model is $Y = {X}\beta$ where $\beta = ({\beta}_1, ..\ , {\beta}_p)$.
Training determines $\beta$ for a given dataset. 
In the context of learned index that uses variable-length string keys, the input to the models is a matrix $X$ with $n$ rows where each row is a numerically encoded key vector of length $p$, and $Y$ (output) is a vector of integer values that represent the keys' positions in the sorted key array.
Even for updating the pre-existing models, the entire $X$ and $Y$ are required to retrain all the models traversed in the hierarchical structure and determine their new $\beta{s}$.
After retraining, the index for a given key can be predicted by performing a series of dot products between the traversed model input $X$ and their corresponding $\beta{s}$.

\niparagraph{Learning the parameters.}
Every $\beta$ can be obtained by inversing the matrix $X$ and multiplying it with the output vector $Y$ (i.e., $\beta = X^{-1}Y$). 
However, computing the inverse matrix $X^{-1}$ can be computationally prohibitive, especially when the matrix size is large.
To tackle the challenge, an existing alternative approach commonly and widely used in practice is to employ a matrix factorization method, known as QR decomposition (QRD) technique.
QRD decomposes a matrix $X$ into a multiplication of two matrices: $Q$, an $n$x$p$-sized matrix with $Q^{T}Q=QQ^{T}=I$, and $R$, a $p$x$p$-sized upper triangular matrix.
Figure~\ref{fig:index_training_process}(b) illustrates the linear algebra operations required to determine $\beta$, leveraging the QRD technique.
At first, the QRD of the input dataset $X$ is performed, which produces $Q$ and $R$. 
After the decomposition, the following operations are performed: (1) computing the inverse of the upper triangular matrix ($R^{-1}$), (2) transposing matrices (${R^{-1}}^{T}$ and $X^{T}$), (3) multiplying the resulting small matrices ($R^{-1}{R^{-1}}^{T}$), and (4) matrix-vector multiplication ($X^{T}Y$).
Note that during training, only the $R$ matrix is required.
%

%QR decomposition is the most compute-intensive mathematical operation of above described training, hence is the acceleration target of \lia. 

\begin{algorithm}[t]
\fontsize{9}{8.5}\selectfont
\LinesNumbered
\tline
\vspace{2pt}
    \SetKwInOut{Input}{Input \quad}	
	\SetKwInOut{Output}{Output \quad}

    \SetKwProg{Fn}{Function}{}{}
	\SetKwFunction{AssignWavelet}{AssignWavelet}
 
    \Input {
        $X$:\ \ Matrix of size $m$ $\times$ $n$ \\
    }
    
    \Output {
        $R$:\ \ Upper triangular matrix of size $n$ $\times$ $n$ \\
    }
    \vspace{1.5ex}
    \For {($i \gets 0$ to $n-2$)}
    {
        $col_{i}$ = $X[i:m,i]$ \\
        $d$ = $\sqrt{\mathbf{dot}\ (col_{i}, col_{i})}$ \\
        $ref_i$ = $\mathbf{cal\_reflector}$\ ($col_{i}$,\ $d$) \\
        %$rfl_{i}[0]$ += $rfl_{i}[0] > 0$ ? $d$ : $-d$ \\
        $\gamma$ = $-2$ $/$ $\mathbf{dot}$($ref_i$, $ref_i$) \\
        \For {($j \gets i$ to $n-1$)}
        {
            $col_{j}$ = $X[i:m,j]$ \\
            $\alpha$ = $\gamma$ $\times$ $\mathbf{dot}$\ ($ref_i$, $col_j$) \\
            $col_j$ = $\mathbf{axpy}$\ ($\alpha$, $ref_i$, $col_j$) \\
            $R[i,j]$ = $X[i,j]$
        }
    }
%    $R[n-1:n-1] = X[n-1:n-1]$
\vspace{1ex}
\bline
\caption{\textbf{Householder QR decomposition.}}
\label{alg:qr-algorithm}
\end{algorithm}

\niparagraph{Householder QR decomposition.}
QR decomposition can be computed using various algorithmic methods~\cite{gram-schmidt, householder, givens-rotation}.
Among these methods, we base \lia on the Householder algorithm~\cite{householder} owing to its relatively enhanced numerical stability, while \lia remains compatible with other alternatives due to their algorithmically similar traits.
Algorithm~\ref{alg:qr-algorithm} illustrates the Householder algorithm~\cite{householder}.
The algorithm has two loops. 
In the outer loop, the algorithm iterates over the columns of the input matrix and calculates a vector, called a reflector ($ref_i$), and a scalar value $\gamma$. 
For each column, the inner loop visits all the columns located on the right of the current column one by one, and updates the visiting column while producing the $R[i][j]$ values.
The nature of this process is fundamentally serial.
%
% It is worth noting that while we select the Householder method, other algorithms exhibit similar traits where they are structured in two loops, with an intrinsic requirement for serial execution.
%

\begin{figure}[t]
        \centering
        \includegraphics[width=0.9\linewidth]{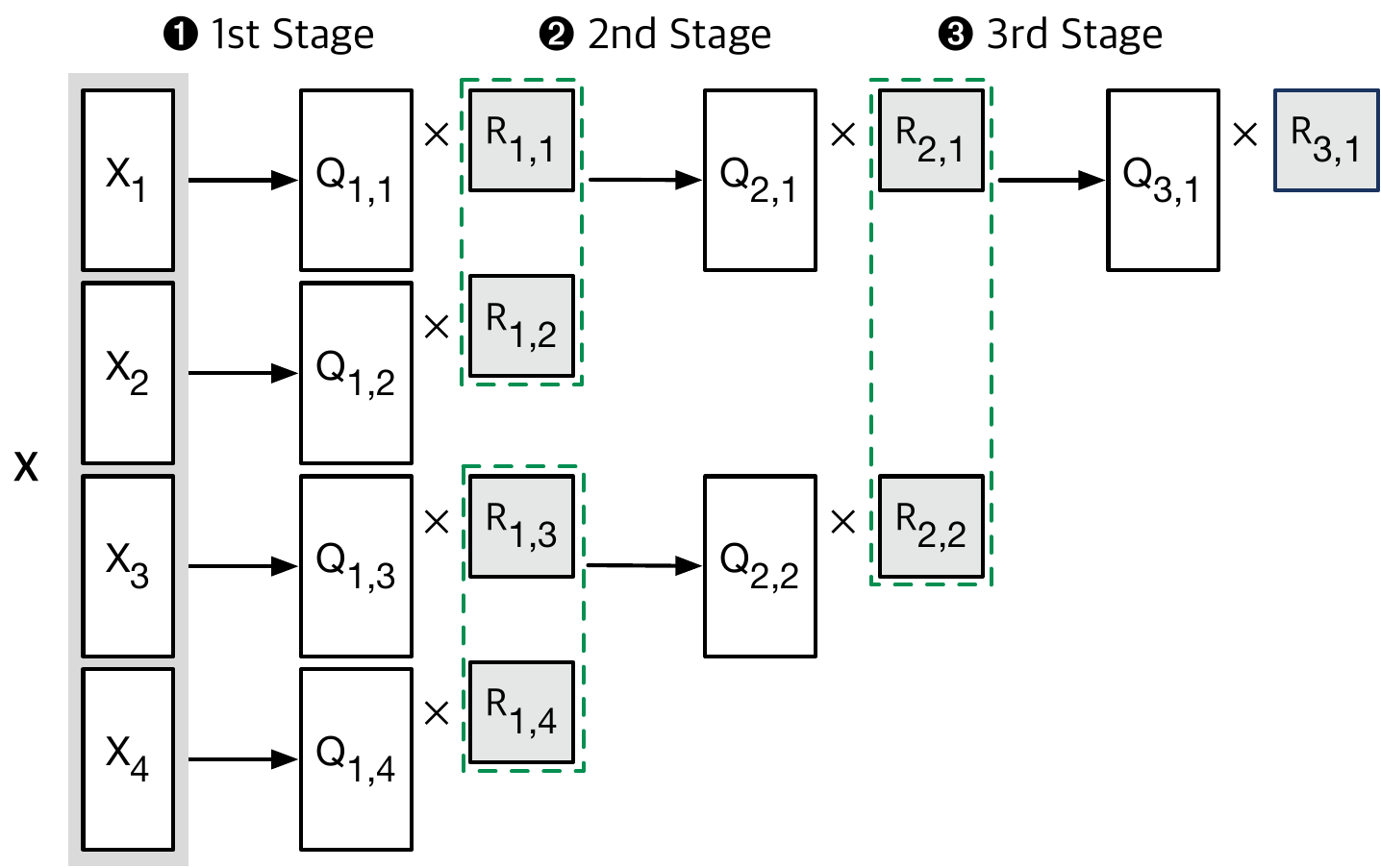}
        \vspace{-2ex}
        \caption{Parallel QR decomposition.}
        \label{fig:parallelized-qrd}
        \vspace{-3ex}
\end{figure}

\niparagraph{Parallelizing QR decomposition.} 
Vanilla QRD algorithms, including Algorithm~\ref{alg:qr-algorithm}, execute sequentially by sweeping through the columns of an input matrix and gradually filling the rows and columns of the $Q$ and $R$ matrices, respectively.
Thus, QRD can be slow for large matrices, as is the case with learned index.
As the number of keys grows, the height of the key matrix $X$ also increases ($n\times{p}$ matrix where $n$ >> $p$), making it a \emph{tall-and-skinny} matrix.
Prior works~\cite{doi:10.1137/1032002, plemmons1988parallel, parallel-qrd} offer a parallelization mechanism customized for tall-and-skinny matrices.
The parallelization mechanism exploits a mathematical property of orthogonal matrix $Q$ that its transpose is equal to its inverse matrix, as depicted with an example in Figure~\ref{fig:parallelized-qrd}.
Let $X$ be an input matrix for an LR model within the tree.
$X$ is decomposed through three steps: 
(1) $X$ is vertically split into smaller sub-matrices ($X_1, X_2, X_3, X_4$) and decomposed into QR matrices in parallel;
(2) the QR decomposition is performed on the vertically concatenated $R$ matrices (\textsf{concat}($R_{1,1}, R_{1,2}$) and \textsf{concat}($R_{1,3}, R_{1,4}$));
(3) finally, the last QR decomposition is applied over \textsf{concat}($R_{2,1}$, $R_{2,2}$) to produce $R_{3,1}$. 
The resulting $R_{3,1}$ is mathematically equivalent to $R$, obtainable by decomposing the $X$ as a whole without parallelization.
\begin{figure}[t]
        \centering
        \includegraphics[width=0.9\linewidth]{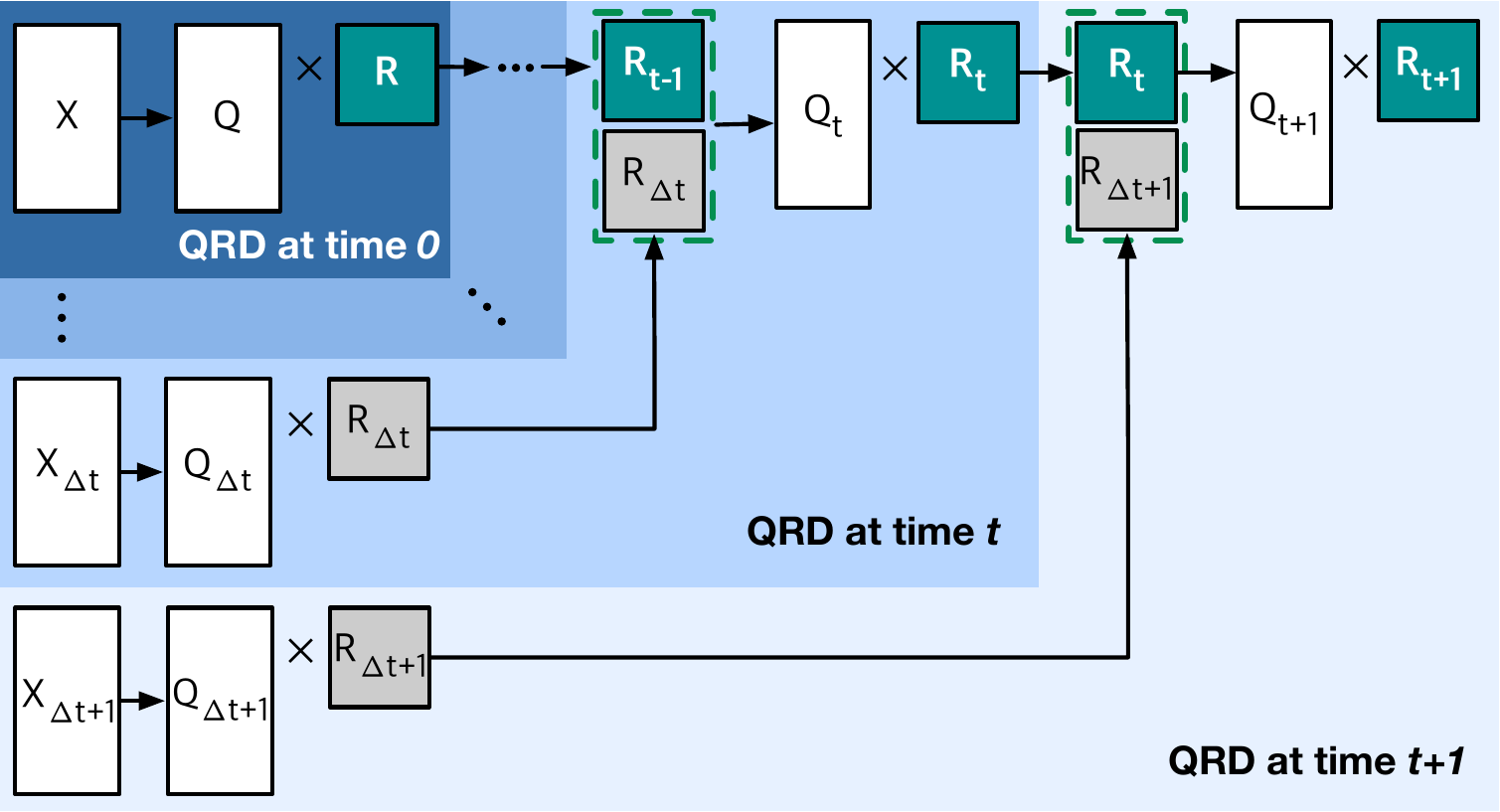}
        \vspace{-1ex}
        \caption{Memoized QR decomposition.}
        \vspace{-2ex}
        \label{fig:incremental-qrd}
\end{figure}
\subsection{\liatitle's Incremental Index Learning} 
\label{sec:lia-algorithm}

\niparagraph{Memoized QRD via computation reuse.}
Exploiting the mathematical insight of parallelized QRD, we modify the vanilla QRD that incurs a heavy amount of computation and devise a memoized QRD.
Figure~\ref{fig:incremental-qrd} shows the memoized QRD algorithm. 
We exclusively consider the case that the number of keys \emph{grows} due to the \texttt{insert} queries\footnote{We will discuss the \texttt{delete} query handling in Section~\ref{sec:lazy-deletion}.}. 
When a learned index is retrained, we require the $R$ matrix corresponding to the current $X$. 
To do so, we memoize the computed $R$ matrix in memory at every retraining invocation ($R_{t}$).
When a retraining is invoked, the rows of collected additional keys $X_{{\Delta}t+1}$ is decomposed. 
Then, similar to the parallelized QRD, we concatenate the $R_{t}$ and $R_{{\Delta}t+1}$, and perform one more QRD to obtain the final $R_{t+1}$. 
Now, $R_{t+1}$ is used for linear model training and cached in memory for the next retraining run. 
Note that \lia's QRD algorithm involves only two small QR decompositions, which significantly reduces the compute load by reusing the performed computations. 
Moreover, the size of each $R$ matrix is $p\times{p}$ where $p$ is the key length, thus is very small and does not incur large memory footprint overhead.
For instance, with a key length of 96, the size of $R$ is merely 72 KB (=96$\times$96$\times$8). 

\niparagraph{\lia's incremental index learning algorithm.}
\lia's incremental index learning algorithm uses the memoized QRD to train the models in the updatable learned indexes.
\begin{algorithm}[t]
\fontsize{8}{8.5}\selectfont
\LinesNumbered
\tline
\vspace{2pt}
	%\caption{\textsf{\textbf{\scriptsize Minimum-Communication Data and Operation Mapping.}}\label{alg:map}}

    \SetKwInOut{Input}{Input \quad}	
	\SetKwInOut{Output}{Output \quad}

    \SetKwProg{Fn}{Function}{}{}
	\SetKwFunction{AssignWavelet}{AssignWavelet}
 
    \Input {
        %$M$[1..n]: List of linear models in learned index \\
        $M_{old}$:\ \ Current linear models \\
        \ $X_{old}$:\ \ \ Current key matrices \\
        \ $X_{\Delta}$: \ \ \ \ Newly inserted key matrices \\
        \ $Y_{old}$:\ \ \ Current index vectors \\
        \ $R_{old}$:\ \ \  Memoized R matrices \\
    }
    
    \Output {
        $M_{new}$:\ \ Updated linear models \\
        \ $X_{new}$:\ \ \ Updated key matrices \\
        \ $Y_{new}$:\ \ \ Updated index vectors \\
        \ $R_{new}$:\ \ \ Newly memoized R matrices \\
    }
    Initialize $M_{new}$ $\leftarrow$ $\emptyset$, $X_{new}$ $\leftarrow$ $\emptyset$, $R_{new}$ $\leftarrow$ $\emptyset$ \\
    %Initialize $X_{old}$ $\leftarrow$ $K_{old}$, $X_{\Delta}$ $\leftarrow$ $K_{\Delta}$, $Y$ $\leftarrow$ $I_{new}$ \\
    \While {($m$ $\in$\ $M_{old}$)}
    {
        $mid$ $\leftarrow$ $m.model\_id$ \\
        \vspace{1ex}
        $X_{new}[mid]$ $\leftarrow$ \textsf{concat}\ ($X_{old}[mid]$, $X_{\Delta}[mid]$) \\
        $Y_{new}[mid]$ $\leftarrow$ \textsf{calc\_index}($Y_{old}[mid]$, $X_{new}[mid]$) \\
        $tmp$ = $(X_{new}[mid])^{T}$ $\times$ $Y_{new}[mid]$ \\
        \vspace{1ex}
        $R_{\Delta}$ $\leftarrow$ $\mathbf{QR}$($X_{\Delta}[mid]$) \\
        $R_{tmp}$ $\leftarrow$ \textsf{concat}($R_{old}[mid]$, $R_{\Delta}$) \\
        $R_{new}[mid]$ $\leftarrow$ $\mathbf{QR}$($R_{tmp}$) \\
        \vspace{1ex}
        $\beta$ = ($(R_{new}[mid])^{-1}$ $\times$ $((R_{new}[mid])^{-1})^{T}$) $\times$ $tmp$\\
        % $R_{inv_T}$ = $(R_{inv})^{T}$ \\
        % $R_{inv,inv_T}$ = $R_{inv}$ $\times$ $R_{inv_T}$ \\
        % $\beta$ = $R_{inv,inv_T}$ $\times$ $X^{T}y$ \\
        \vspace{1ex}
        $M_{new}$$[mid].$$\beta$ $\leftarrow$ $\beta$ \\
        % $K_{new}$[mid] $\leftarrow$ $X_{new}$[mid] \\
        % $X_{new}$[mid].append($X_{new}$) \\
        % $R_{new}$[mid].append($R_{new}$) \\
    }
\vspace{1ex}
\bline
\caption{\textbf{Incremental index learning algorithm.}}
\label{alg:incremental-index-learning}
\end{algorithm}
Algorithm~\ref{alg:incremental-index-learning} describes \lia's training process. 
The algorithm loops over the list of linear models in the hierarchical structure, which need to be updated.
It concatenates the existing keys $X_{old}$ with new keys $X_{\Delta}$ to obtain $X_{new}$, calculates indexes for the new keys to update $Y_{old}$ with $Y_{new}$, and computes the $(X_{new})^{T}Y_{new}$. 
Then, the algorithm performs the memoized QRD, which results in $R_{new}$.
%
% Given the $R_{new}$, the rest of linear algebraic operations are performed to obtain the $\beta$.
Using $R_{new}$, the algorithm obtains the $\beta$ and updates the model parameters with the new $\beta$.
The obtained $R_{new}$ is memoized for next retrainings. 
The same training process is repeated until the models of all leaf and internal nodes in the index structure are updated.

\begin{figure}[t]
        \centering
        \includegraphics[width=0.75\linewidth]{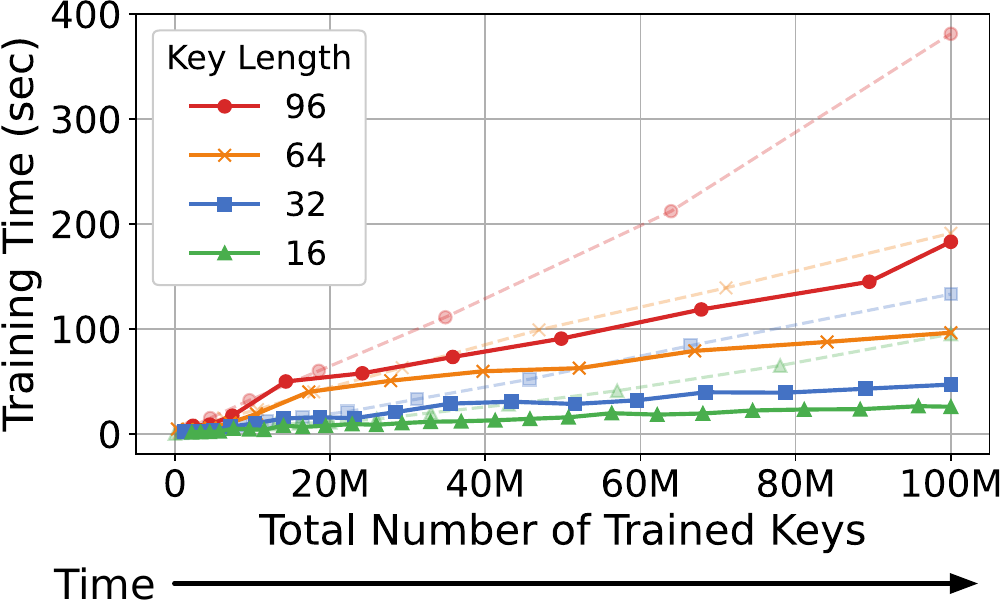}
        \vspace{-2ex}
        \caption{Increasing retraining time as the total number of keys increases with CPU-based memoized QRD on SIndex. Markers on the same line represent sequential retraining runs, where markers positioned to the left precede those on the right. Key lengths are 16, 32, 64, and 96. For comparison, the shaded lines depict the results presented in Figure~\ref{fig:scalability}.\vspace{0ex}}
        \label{fig:incremental-sindex-scalability}
        \vspace{-2ex}
\end{figure}

\begin{figure}[t]
        \centering
        \includegraphics[width=1.0\linewidth]{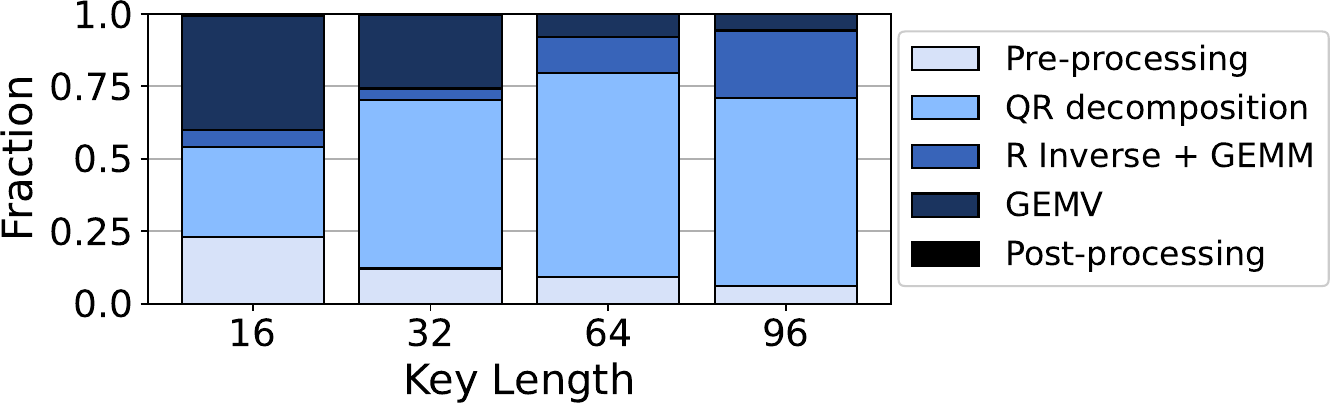}
        \vspace{-5ex}
        \caption{Breakdown of linear model training runtime.}
        \label{fig:linear-regression-training-breakdown}
        \vspace{-2ex}
\end{figure}

\begin{figure*}[t]
        \centering
        \includegraphics[width=0.93\linewidth]{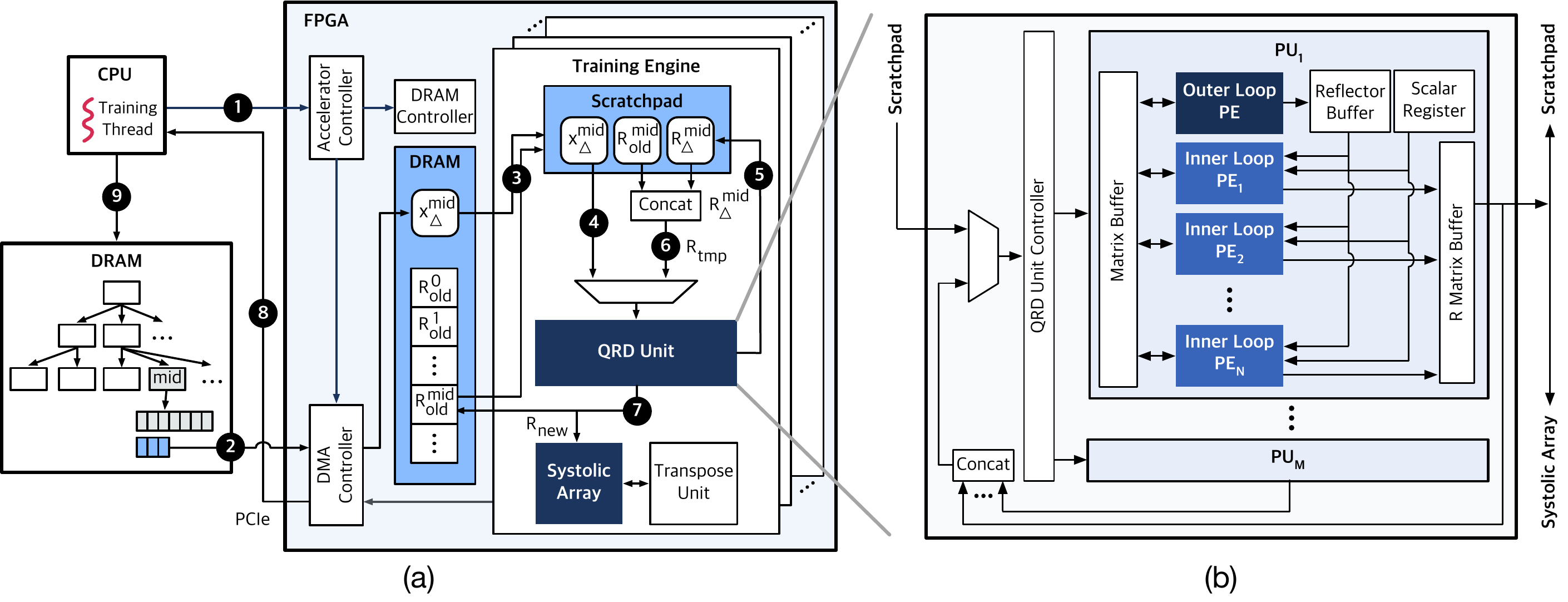}
        \vspace{-2ex}
        \caption{(a) \lia's system where the CPU runs the training thread to issue jobs to the accelerator. Accelerator executes the training operations on the training engine; (b) Microarchitecture of QR decomposition unit.}
        \vspace{-1ex}
        \label{fig:system}
\end{figure*}

\niparagraph{Limitation of software-only solution.}
We observe that the \lia's learning algorithm already substantially reduces the computational cost of training the learned index, even when implemented in software without hardware acceleration.
Figure~\ref{fig:incremental-sindex-scalability} shows the improved training time with the proposed memoized algorithm.
Compared to the baseline reported in Figure~\ref{fig:scalability}, which is presented as dimmed lines in Figure~\ref{fig:incremental-sindex-scalability}, the retraining time is reduced for all the evaluated key lengths, as reflected by the slopes of line graphs. 
Moreover, this shortens the retraining interval, as shown in Figure~\ref{fig:incremental-sindex-scalability} reporting a greater number of data points (markers), each corresponding to a retraining. 
However, Figure~\ref{fig:incremental-sindex-scalability} also shows that the resulting reduction in training time is insufficient, still extending up to 200s.
\niparagraph{Acceleration target determination.}
This observation motivates us to devise an efficient and performant hardware accelerator for training. 
However, the first crucial step is to determine the acceleration targets for offloading to the hardware.  
For this purpose, we first characterize the core compute kernels of training.
Figure~\ref{fig:linear-regression-training-breakdown} shows the results as we vary the key length from 16 to 96. 
We look into four kernels: (1) training data matricization, (2) QR decomposition, (3) R matrix inverse calculation and matrix-matrix multiplication, and (4) matrix-vector multiplication.
As the data matricization is mostly memory copy, it needs to be performed by CPU. 
We also rule out matrix-vector multiplication from the acceleration targets since it requires a memory copy for the entire $X$ matrix from host to FPGA.
To this end, this work focuses on accelerating QR decomposition, R inverse, and GEMM operations on the FPGA.

\section{\liatitle System Design}
\label{sec:system}

While \lia employs the incremental learning algorithm to reduce the computation load, we enhance this algorithmic approach by incorporating an FPGA accelerator and customized runtime software to further accelerate \lia's training.
We first describe the overview of \lia's system, and then, elaborate each component in detail.

\subsection{FPGA-Accelerated Training Infrastructure}
FPGAs have been commonly used as a successful platforms for acceleration~\cite{tabla:hpca, cosmic:micro, dnnweaver:micro:2016, dana} and are even deployed in cloud datacenters~\cite{catapult}. 
Figure~\ref{fig:system}(a) depicts the \lia system accelerated using FPGA.
As in existing learned index systems, \lia employs a multi-core CPU that can serve both inference and training. 
However, \lia also comes with an FPGA accelerator to offload training computation. 
We chose FPGA as the acceleration platform owing to its customizability to index-specific algorithms and high energy efficiency, which is crucial for index systems since training computations are consistently conducted throughout their lifespan. 
Unlike the existing systems, \lia only runs a single training thread to not only compute the non-accelerated memory-bound kernels, but also manage the data transfer between host and FPGA and control the accelerator invocations.
The training thread iterates over a list of linear models within the hierarchical structure and initiates the retrainings of these models one by one on available Training Engines (TEs). 
%
% Each linear regression model is trained on a single TE in its entirety. 
%
To train a model, the newly inserted keys accumulated in the model's buffer ($X_{\Delta}$) are first copied from host to FPGA.
FPGA's off-chip memory maintains an array of $R_{old}$ matrices, which are memoized from the previous retraining runs. 
In the figure, the superscript $mid$ on the $R$ and $X$ matrices refers to the model ID. 
After the memory copy from the host to FPGA is completed, the training thread sets a control register in the accelerator controller, scheduling the training computation to an available TE.
The training thread is also responsible for updating the model parameters, which occurs repeatedly during runtime, allowing the index to integrate new keys.
\subsection{Accelerator Architecture}
\niparagraph{Training Engine.}
Figure~\ref{fig:system}(a) also depicts the TE architecture.
The first computation performed by TE is the \lia's QRD algorithm described in Section~\ref{sec:linear-regression-training}.
The TE feeds $X_{\Delta}$ to the QRD unit.
It then obtains the $R_{\Delta}$, which is concatenated with the memoized $R_{old}$ in the scratchpad memory to produce the $R_{tmp}$.
This $R_{tmp}$ is then fed to the QRD unit as an input that produces $R_{new}$.
The next step is to perform $R_{new}$ matrix inversion and matrix-matrix multiplication between the inverse and its transpose.
%
% To efficiently perform the matrix inverse operation, 
We exploit a parallelized matrix inverse algorithm, namely Heller's algorithm~\cite{hellers_algorithm}, which effectively converts a matrix inverse into a series of recursive matrix-matrix multiplications.
As we transform all needed operations into a series of matrix-matrix multiplications, a systolic-array accelerator equipped with a transpose unit can complete all the necessary kernel executions. 
Once the computation is completed, the accelerator controller uses a control flag to inform the training thread about the completion.

\niparagraph{QRD Unit.}
Due to its computational intensity in mathematical problems, QRD has been a target for hardware acceleration~\cite{qrdaccel-fpga18, qrdaccel-fpl12, parallel-qrd}. 
We devise the architecture of our QRD unit inspired by an existing QRD accelerator~\cite{parallel-qrd}, which executes the Householder algorithm described in Algorithm~\ref{alg:qr-algorithm}.
Figure~\ref{fig:system}(b) shows the microarchitecture of the QRD unit in each Training Engine. 
QRD unit constitutes an array of Processing Units (PUs), each of which executes a QRD.
%, leveraging the parallelization approach described in Section~\ref{sec:linear-regression-training}.
%
The results of PUs are concatenated and stored back to the matrix buffer for the next stage of QRD (Figure~\ref{fig:parallelized-qrd}).
Each PU first gets its input data from scratchpad memory ($X_{\Delta}$ or $R_{\Delta}$) and stores them in the matrix buffer. 
Then, the outer loop in Algorithm~\ref{alg:qr-algorithm} is performed at the ``Outer Loop PE'', which calculates the reflector and $\gamma$.
These two inputs are sent to a set of "Inner Loop PEs", which are responsible for calculating $R_{new}[i][j]$ for different columns in parallel.
Each ``Outer Loop PE'' and ``Inner Loop PE'' is equipped with a vector of multiply-and-accumulate (MACC) units for dot products.
The resulting $R_{new}$ matrix is sent to the scratchpad memory and replaces $R_{old}$ for future retrainings.
\vspace{-2ex}

\subsection{Runtime Software Interface}
\label{sec:runtime}

As emphasized in Section~\ref{sec:rmi}, in designing the \lia system, we leverage a commonality of most learned indexes that they use linear regression as their backend machine learning models.
This unique property enables us to build an abstraction between various learned indexes and our hardware accelerator solution. 
Therefore, \sia could be readily adopted by any linear model-based learned index systems.
To transparently develop the abstraction and facilitate the use of underlying acceleration solution, we encapsulate the \lia accelerator along with its device driver and accelerator invocation runtime software as a library.
In fact, as existing learned index systems often employ LAPACK, a famous linear algebra library, we propose \sia's interfaces to be equivalent to the LAPACK's, so that the integration of \sia with the existing systems becomes straightforward.
The runtime interface of \lia includes two functions: (1) \texttt{cold\_train}: a function for full model training with key matrix and key's position vector, and (2) \texttt{incre\_train}: a function for incremental learning with memoization that takes the memoized $R$ matrix as an additional argument.
These two functions closely resemble the LAPACK's \emph{gels} function, enabling existing updatable learned indexes to leverage \lia's incremental index learning algorithm and hardware acceleration with minimal software modifications.

\subsection{Lazy Delete Query Handling} 
\label{sec:lazy-deletion}
While this paper has focused on the \texttt{insert} query handling thus far, updatable learned indexes must be able to handle \texttt{delete} queries as well.
Conventional updatable learned indexes handle these \texttt{delete} queries through retraining, similarly to the \texttt{insert} queries. 
In contrast, our incremental learning algorithm exploits a memoization technique, which relies on the assumption that the existing keys used to compute the memoized $R$ matrix are not changed.
Therefore, the removal of keys from the index inevitably forces \lia to discard the memoized $R$ matrix and necessitates a cold training, which undercuts the advantages of our proposed technique. 
To tackle this problem, we employ a \emph{lazy} delete handling technique where the deleted keys are simply flagged as ``deleted'', yet the information of these deleted keys still remains in the memoized $R$ matrices.
This way, our incremental training method remains effective during retraining.
It is important to note, however, that upon marking as ``deleted'', the key string and associated value data are immediately erased from the indexes for security purposes. 
Memoized R matrices for the ``deleted'' keys are eliminated during cold training, where the models are trained from scratch without utilizing the memoized matrices.
%
%This way, retrainings can still benefit from our incremental training method, while the false accesses to the deleted keys will be correctly handled by checking the ``deleted'' marks and failing the accesses. 
%
%The marked keys are actually eliminated during cold retraining, where the models are trained from scratch without utilizing the memoized matrix.
%
Note that our lazy deletion technique does not affect the functionality of indexes, but only influences performance, since deleted-yet-unremoved information would lower the prediction accuracy and end up increasing the linear search cost for mispredicted accesses.  
However, we observe that lazy deletion has a marginal impact on performance, with less than 5\% overhead (see Section~\ref{sec:lazy_del_eval}).

\subsection{Implication of Node Split and Merge}
\label{sec:split-merge}
The hierarchical structure of learned index necessitates structural modifications as new keys are inserted or deleted.
Hierarchical learned index structures undergo structural modifications through either \emph{split} or \emph{merge}. 
Model \emph{split} involves partitioning the keys assigned to a node into two nodes when the accuracy of the corresponding model drops, while model \emph{merge} combines two nodes into one when both have sufficiently high accuracy.
\lia employs the same threshold determination mechanism for split and merge as the default learned index system, without any modifications. 
Note that \lia should perform cold trainings for split nodes as they lack memoized $R$ matrices, while for merged nodes, \sia can merge the $R$ matrices and use the merged $R$ matrix for further incremental training.  

\section{Evaluation}
\label{sec:evaluation}
%
%To evaluate the effectiveness of \sia, we integrate our solution, \sia, to existing learned indexes: \sindex, \alex, and \lipp.
%}
To evaluate the effectiveness of \lia, we harness two open-source benchmark suites, YCSB and Twitter cache trace, using two real-world datasets, Amazon review and MemeTracker.
We evaluate throughput, system-level energy efficiency, and memory usage of \sia-accelerated learned indexes, compared to other index structures.
%

% \begin{figure}[t]
%         \centering
%         \vspace{-2.5ex}
%         \includegraphics[width=1.0\linewidth]{fig/dataset-cdf.pdf}
%         \vspace{-4.5ex}
%         \caption{Cumulative distribution function (CDF) of keys. (a)-(c) illustrate key distributions of different datasets used for YCSB, while (d)-(g) show the distributions for different Twitter cache clusters.}
%         \label{fig:dataset-cdf}
%         \vspace{-2.5ex}
% \end{figure}
%
\begin{figure*}[t]
        \centering
        \includegraphics[width=0.8\linewidth]{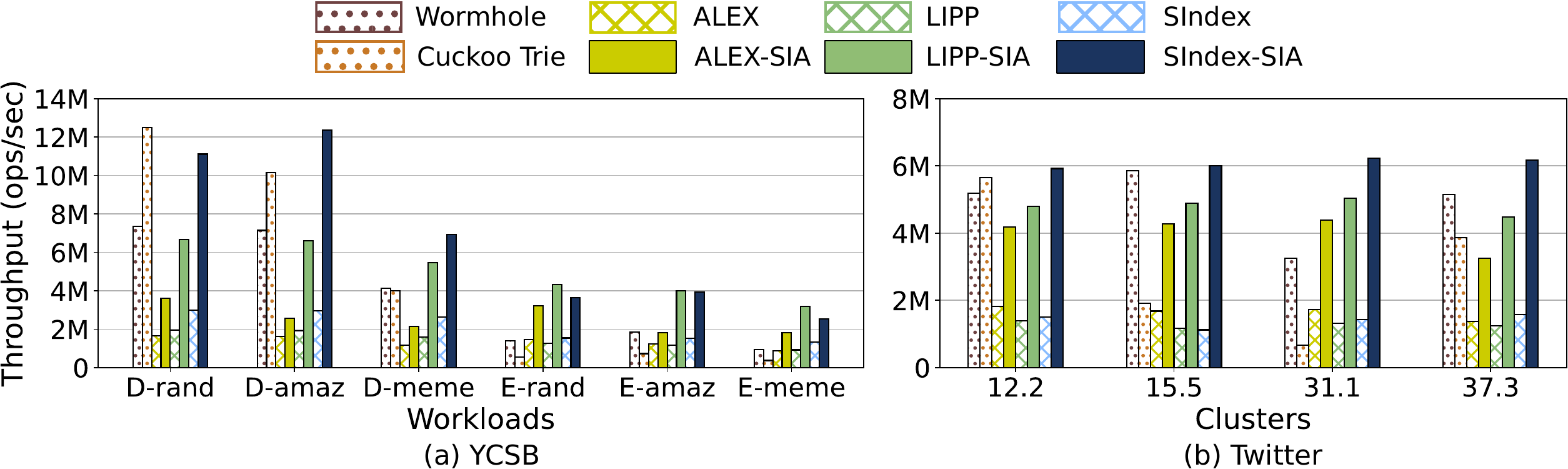}
        \vspace{-2ex}
        \caption{Throughput comparison of \emph{non}-learned (conventional) and learned indexes for YCSB and Twitter cache trace.}
        \label{fig:throughput}
        \vspace{-1ex}
\end{figure*}

\subsection{Methodology}
\label{sec:method}
\niparagraph{YCSB.}
To evaluate \sia, we primarily use a real-world key-value store benchmark suite, YCSB~\cite{ycsb}.
%
%YCSB~\cite{ycsb} is an established benchmark suite renowned for its standardized method to evaluate key-value stores, including learned index systems.
%
YCSB contains six diverse workloads (A-F), each characterized by its unique mix of query types. 
As \sia is for updatable learned index systems, we focus on the two workloads among the six, which include \emph{insert} queries: (D) \emph{read latest} that tend to have \texttt{read} queries for recently inserted keys, along with roughly the 5\% of \texttt{insert} queries, and (E) \emph{short ranges} that consists of 95\% \texttt{range} queries, and 5\% of \texttt{insert} queries.
%
% (E) \emph{short ranges} that mostly constitutes \texttt{range} queries which can be performed with two \texttt{read} queries, and 5\% of \texttt{insert} queries.
%
Note that while YCSB's query compositions mirror real-world application patterns, the key lengths do not.
To better emulate real-world key-value stores, we employ two genuine string datasets: Amazon review data and the MemeTracker dataset.
Amazon review data (\emph{amaz})~\cite{amazon-review-data} is collected from user reviews on products from Amazon with the user IDs as keys of length 12.
MemeTracker dataset (\emph{meme})~\cite{memetracker} comprises quotes and phrases collected from the web and online news URLs referring to them with the URLs as keys of length 128.
We use these datasets since they are widely used in prior works~\cite{wormhole, treeline-22, sindex} to evaluate the string-key key-value stores. 
Additionally, we use a randomly synthesized dataset (\emph{rand}) with a uniform key distribution.

\niparagraph{Twitter cache trace.}
Complementing YCSB, we also utilize the Twitter cache trace~\cite{twitter-cache-trace} to enrich our experimental methodology. 
Twitter cache trace constitutes a pile of indexing traces collected from Twitter clusters, which allows it to concurrently serve as a workload and a dataset.
Among the provided 54 cluster traces, we specifically select four cluster traces with a relatively significant volume of update queries, each of which exhibits a distinct query composition, represented by the following tuples of (cluster ID, update query ratio): (12.2, 43\%), (15.5, 59\%), (31.1, 56\%), (37.3, 42\%).
%
% Figure~\ref{fig:dataset-cdf} reports the varying key distributions of different datasets that allow us to effectively examine the diverse behaviors of updatable string-key learned indexes. 

\niparagraph{Baselines.}
As baselines, we use three state-of-the-art learned indexes, ALEX~\cite{alex}, LIPP~\cite{lipp}, and SIndex~\cite{sindex}, all of which are chosen for their open-source implementations available at our disposal.
We added the variable-length string key and multi-threading support on top of ALEX and LIPP, as they lack the features.
We built their corresponding \sia-accelerated counterparts by integrating the implementation with our \sia library.
%
%As the open-source implementations of ALEX and LIPP lack support for string keys and concurrency, we modified the implementations such that they can handle variable-length string keys and support multi-threading using the concurrency mechanisms employed by XIndex~\cite{xindex} and SIndex~\cite{sindex}.
%
%
Note that while the three systems have disparities in how to initially build the indexes through bulk loading (e.g., top-down vs. bottom-up), it does not affect our performance evaluations because the index building only requires cold retrainings, which cannot exploit the proposed incremental index learning algorithm. 
Furthermore, we include comparisons between \sia-accelerated learned indexes and two state-of-the-art \emph{non}-learned indexes, Wormhole~\cite{wormhole} and Cuckoo Trie~\cite{cuckoo-trie}, all of which support variable-length string keys.
Wormhole~\cite{wormhole} is an optimized B-tree in which part of the tree is replaced with a trie utilizing hashes.
Cuckoo Trie~\cite{cuckoo-trie} is a hash-based trie index that achieves high performance through overlapping memory accesses.
%
%Additionally, we use a GPU-accelerated variant of SIndex, called \sindexgpu, that offloads the model training computation to GPU.
%
%
%
%We compare \lia with two state-of-the-art \emph{non}-learned indexes, Wormhole~\cite{wormhole} and Cuckoo Trie~\cite{cuckoo-trie}, and one updatable learned index, SIndex~\cite{sindex}, all of which support variable-length string keys.
%
%\wormhole~\cite{wormhole} is an optimized B-tree in which part of the tree is replaced with a trie utilizing hashes.
%
%\cuckootrie~\cite{cuckoo-trie} is a hash-based trie index that achieves high performance through overlapping memory accesses.
%
%Additionally, we use \sindex~\cite{sindex} and its GPU-accelerated variant, called \sindexgpu, that offloads the model training computation to GPU. 
%

\niparagraph{System specifications.}
The \sia-accelerated learned index systems are equipped with a 16-core Intel Xeon Gold 6226R and 128 GB DRAM.
%
% To facilitate the analysis, we turn off the hyper-threading. 
%
For building \sindexgpu, GPU-accelerated variant of SIndex, we employ NVIDIA GeForce RTX 2080 TI GPU along with the same CPU and memory configuration.
\sindexgpu uses CuSolver library in CUDA version 11.7.
%
% The systems are deployed on Ubuntu 18.04.4 LTS with the kernel version 4.15.0-175. 
%
For the runtime measurement of the baseline learned index systems, we use a highly-optimized, parallel linear algebra library, Intel oneAPI Math Kernel Library (MKL) 2019.0. 

\begin{table}[t]
        \centering
        \caption{Hardware specifications and resource utilization of the Intel Arria 10 with the configuration of with 4 TEs, each having 2 PUs, and each PU containing 3 Outer Loop PEs.}
        \vspace{-3ex}
        \includegraphics[width=1.0\linewidth]{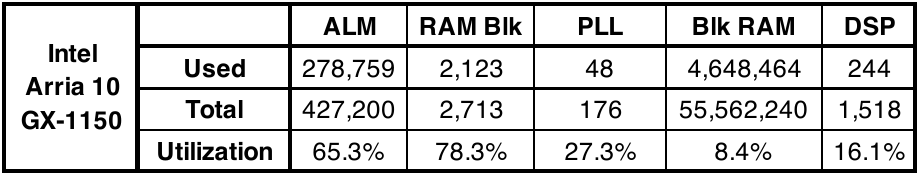}
        \label{tab:fpga}
        \vspace{-5ex}
\end{table}

\niparagraph{FPGA platform details.}
Table~\ref{tab:fpga} shows the hardware resource specification of the evaluated FPGA platform, Intel Arria 10 GX-1150, and its utilization when we program our accelerator on it. 
We develop a custom accelerator controller on the programmable logic to interface with the device's main memory. 
We synthesize the hardware with Quartus II v20.1, and achieve a frequency of 272 MHz.
%

%
% \niparagraph{Ideal system.}
% %
% Throughout the evaluation, we also compare \lia against an ideal system that serves updated keys, without necessitating any training time for their incorporation.
% %
% This system does not use CPU cycles to manage the training thread and the accelerator, which is the fictitious case where \lia tries to closely approach. 
%

\niparagraph{Power measurement.}
To measure the end-to-end system power, we use an off-the-shelf power meter, WATTMAN HPM-100A~\cite{wattman-power-meter}.
This power meter is placed between the power outlets and the servers, which are configured with various processor combinations, including CPU-only, CPU-GPU, and CPU-FPGA setups.
The measured system power can be monitored per each second through the vendor-provided software, which we average over the experiment runtime.
%

%While the improved throughput closely approaches to the throughput of idealized system, there is still a room that cannot be obtained through our system, because unlike ours, the idealized system does not have any keys waiting for training and being absorbed to the model in the buffer at any point of time during the runtime, and there is also 

%
\subsection{Experimental Results} 
\label{sec:results}
\subsubsection{Throughput}
Figure~\ref{fig:throughput} shows the throughput comparison results among two \emph{non}-learned indexes (\wormhole and \cuckootrie), three learned indexes (\alex, \lipp, \sindex), and their \sia-accelerated counterparts (\alexsia, \lippsia, \sindexsia).
%
%Figure~\ref{fig:throughput} shows the throughput comparison results among two \emph{non}-learned indexes (\wormhole and \cuckootrie), SIndex (\sindex), \lia only with the memoization-based incremental learning technique in software (\liasw), the hardware accelerated \lia (\liahw), and the idealized system (\liaideal).
%

%
\niparagraph{YCSB results.}
Figure~\ref{fig:throughput}(a) illustrates the results using two YCSB workloads across three datasets: \emph{rand}, \emph{amaz}, and \emph{meme}.
Although there is some variability in the results, we observe a consistent trend that the \sia-accelerated indexes outperform the learned index baselines, as well as the conventional, \emph{non}-learned index baselines.
%
%Although there is some variability in the results, we observe a consistent trend that the \sia-accelerated indexes outperform the conventional, \emph{non}-learned index baselines. 
%
This translates to approximately an average 2.6$\times$ throughput improvement over CPU-only learned index systems.
This substantial enhancement is attributed to \sia's utilization of both iterative learning algorithm and customized hardware accelerator.
%
% These components effectively reduce computational load and harness the FPGA accelerator to deliver high performance training.
%
This approach dedicates the majority of CPU cores to inferences, while the system allocates only one training thread for memory-bound kernels and accelerator management, not performing any expensive 
% linear algebraic 
operations.

\begin{figure}[t]
        \centering
        \includegraphics[width=0.8\linewidth]{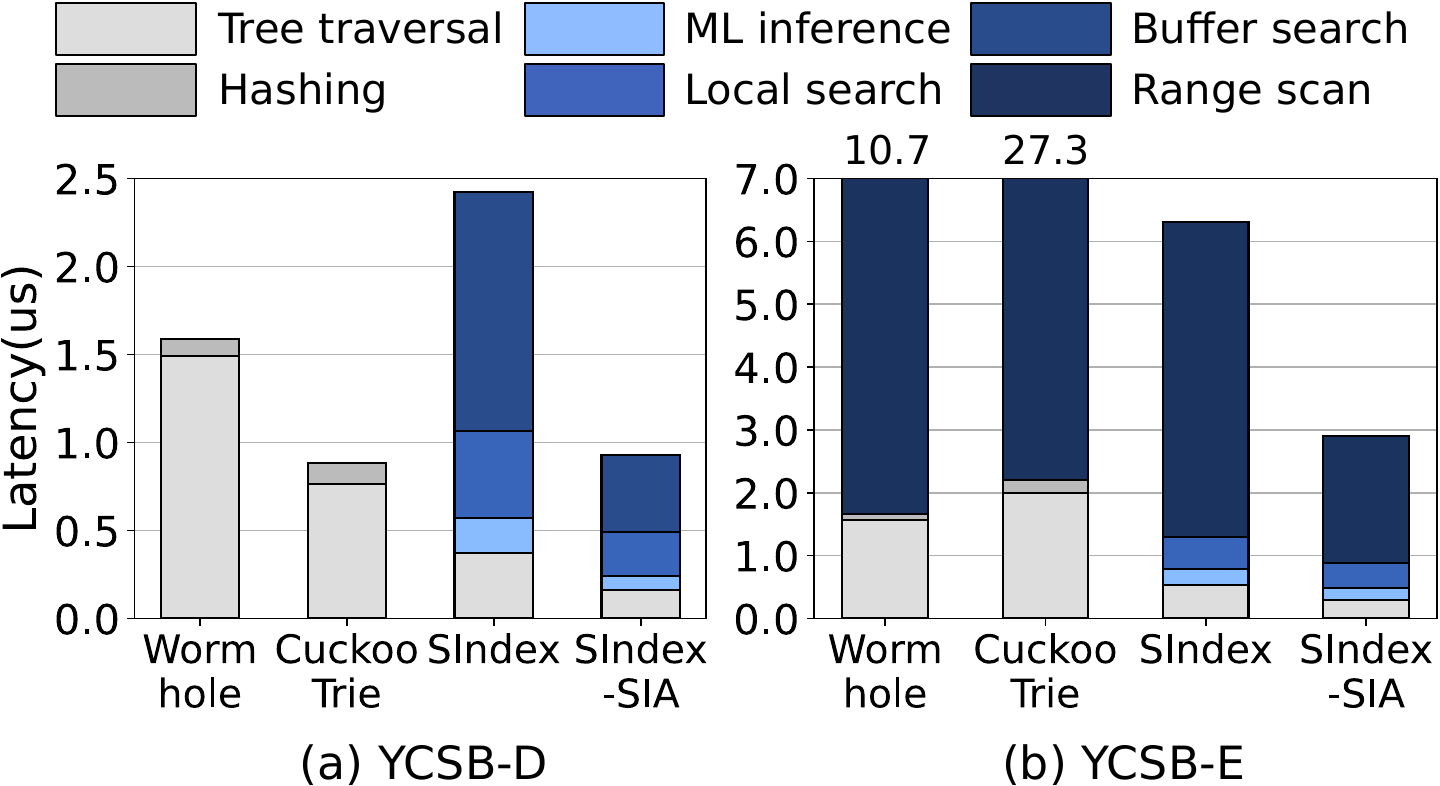}
        \vspace{-2ex}
        \caption{Latency breakdown for YCSB D/E workloads using \emph{rand} dataset. Range scan includes buffer search for YCSB-E.}
        \label{fig:latency-breakdown}
        \vspace{-2ex}
\end{figure}

\niparagraph{Twitter cache trace results.}
Figure~\ref{fig:throughput}(b) reports the throughput results for Twitter cache trace.
Twitter cache trace has diverse key lengths that range from 19 to 82.
As the key length directly affects the computational load, there are variations among clusters in the throughput results.
On average, the \sia-accelerated learned indexes offer 3.4$\times$ throughput improvement over CPU-only systems, representing a more substantial performance improvement than observed in the YCSB scenario. 
The larger gain comes from that the dataset of Twitter cache trace has generally longer keys, making the key matrix larger, which can be better parallelized by the accelerator. 
Overall, the results suggest that \sia is an effective solution for enabling updatable string-key learned indexes without suffering from performance bottlenecks caused by training computations.
%
% One notable observation is that the keys in the Twitter cache trace have common substrings.
% %
% \cuckootrie leverages the common substrings to reduce the index memory usage, while they observe no throughput improvement from the reduction.
% %
% \wormhole is agnostic to the common substrings and reports that it performs worse for the keys with common substrings since it negatively affects their hash efficiency.
% %
% In contrast, \sindex omits common prefixes and only trains unique substrings, which leads to smaller training overhead and better performance.
%
% In summary, the results demonstrate that \lia better leverages the value characteristic of variable-length string keys and achieves an additional performance boost.
%
%Overall, the results suggest that \sia is an effective solution for enabling updatable string-key learned indexes without suffering from performance bottlenecks caused by training computations.

%
\subsubsection{Query Latency}
%
%Latency is another essential metric to evaluate the capability of an indexing system.
%
%Figure~\ref{fig:latency-result} shows the latency measurement result for YCSB and Twitter cache trace.
%
%In these experiments, we observe a similar trend to the throughput results where learned indexes accelerated with \sia offer competitive or superior performance compared to the baseline indexes.
%
To understand the source of performance improvements, we further analyze the query latency for YCSB workload (D) and (E), and present the breakdown results in Figure~\ref{fig:latency-breakdown}.
Non-learned indexes, \wormhole and \cuckootrie, require traversal through their tree structures, which often involve multiple DRAM accesses, leading to high query latency.
In contrast, learned indexes (\sindex and \sindexsia) require much fewer memory accesses for graph traversal.
In fact, the depth of hierarchical learned index structure of \sindex is only two, which imposes significantly lower memory access overhead than the alternatives.
%
%In fact, the depth of hierarchical learned index structure in all of our experiments is only two, which imposes significantly lower memory access overhead than the alternatives. 
%
As the cost of these benefits, the learned indexes must pay other costs such as \textsf{\normalsize{ML inference}}, \textsf{\normalsize{local search}} in case of misprediction, and \textsf{\normalsize{buffer search}} for seeking the ``not-yet-trained'' keys. 
The outcomes of the study reveal that the \textsf{\normalsize{buffer search}} is the largest overhead, especially for \sindex, because it piles up a large number of keys in the buffer due to slow retraining. 
On the contrary, \sia accelerates the retrainings and frequently empties the buffers of \sindexsia, which substantially reduces the buffer search latency, directly leading to the total latency reduction.

\begin{figure}[t]
        \centering
        \includegraphics[width=0.85\linewidth]{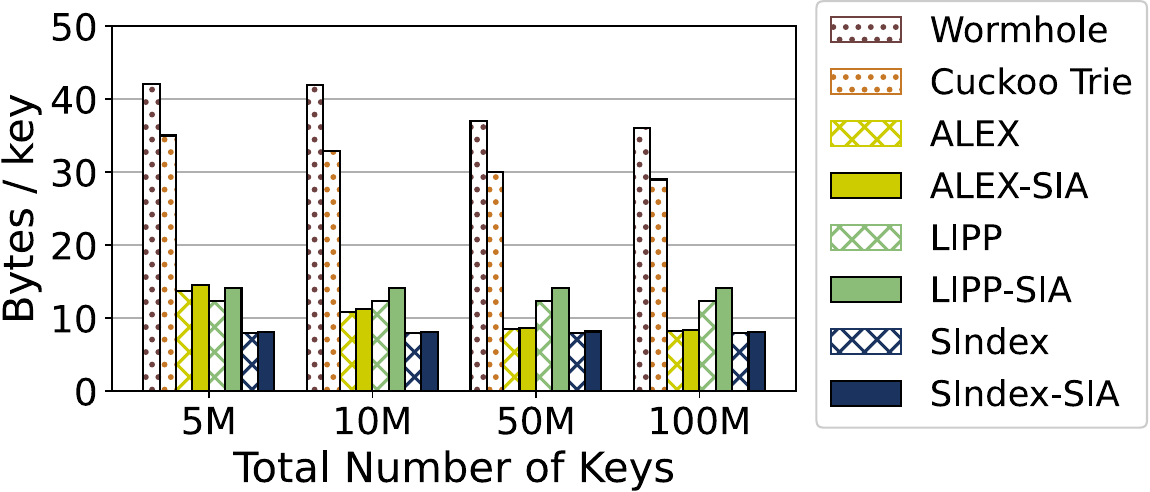}
        \vspace{-2ex}
        \caption{Memory consumption of traditional (non-learned) indexes, baseline learned indexes, and learned indexes with \sia. Key and value data is excluded.}
        \vspace{-2ex}
        \label{fig:memory-usage-comparison}
\end{figure}

\begin{figure}[t]
        \centering
        \includegraphics[width=1.0\linewidth]{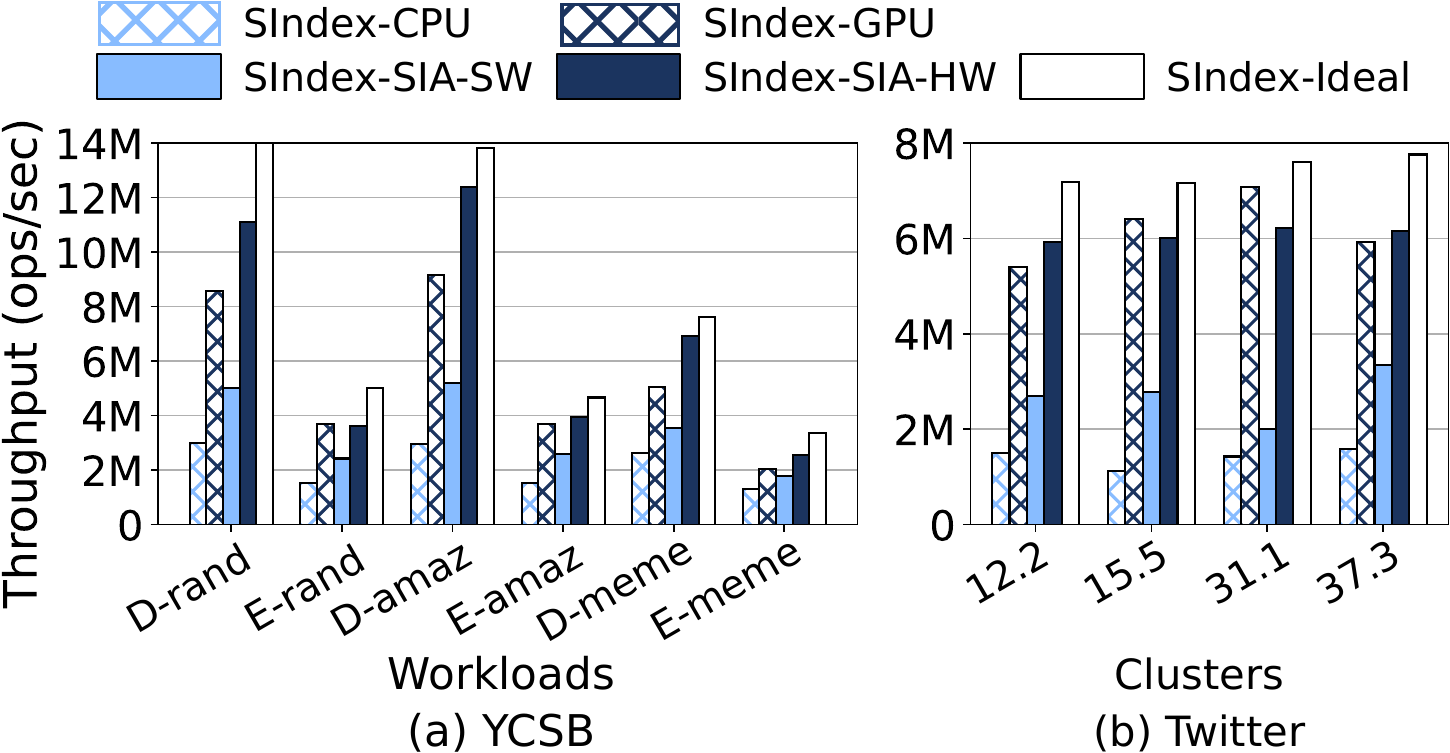}
        \vspace{-5ex}
        \caption{Ablation study results using \sindex variants.}
        \label{fig:ablation-study}
        \vspace{-3ex}
\end{figure}

\subsubsection{Memory Usage}
Figure~\ref{fig:memory-usage-comparison} reports the memory usage of five baselines (learned and \emph{non}-learned) and three \sia-accelerated learned indexes.
To specifically assess the memory usage difference among the indexes, we exclusively measure the memory usage for indexes, not key and values.
%
% As the number of keys increases, the memory usage for \sindex and \lipp variants remains constant since learned indexes effectively squeeze the information stored in hierarchical data structures into a set of lightweight machine learning models.
%
% Conversely, \alex and \alexsia adopt a ``gap'' mechanism to deal with inserted keys, resulting in the allocation of more memory than needed for the currently existing keys.
%
% Consequently, the memory usage per key decreases as the number of keys increases since the overhead for gap remains the same.
%
Learned indexes typically require significantly less memory because they efficiently compress the key-position mapping information from hierarchical data structures into a series of compact machine learning models.
\sia incurs marginal overhead in memory usage as it must additionally store the $R$ matrices for memoized computation.
However, the average overhead measured in our experiments is merely 6.0\%, which is negligible and justifiable with the significant performance improvements.
%
%Figure~\ref{fig:memory-usage-comparison} reports the memory usage of \lia and the three baselines.
%
%To isolate the memory usage difference among the indexes, we exclusively measure the memory footprint occupied by the index structure, not the key and values.
%
%As the number of keys increases, the memory usage for \sindex and \lia remains constant since learned indexes effectively squeeze the information stored in hierarchical data structures into a set of lightweight machine learning models.
%
%Note that \lia imposes overhead in the memory usage since it must additionally store the $R$ matrices for memoized computation. 
%
%However, when we measure the overhead in our experiments, it is only on average 1.4\%, which is negligible and justifiable with the significant performance improvements. 

%
\subsubsection{Ablation Study}
For a more thorough analysis of the factors contributing to performance improvements, we focus on the \sia-accelerated \sindex and conduct an ablation study.
Figure~\ref{fig:ablation-study} compares the throughput of the five \sindex variants.
\sindexgpu is a system offloading retraining computation to GPU, while \siaideal is a system equipped with an infinitely fast accelerator that trains models in zero time.
On the other hand, \siasw and \siahw are the \sia-accelerated \sindex systems with algorithm-only and algorithm-hardware co-designed \sia solutions, respectively.
\sindexgpu achieves 2.3$\times$ throughput improvement compared to the default CPU baseline, \sindexcpu. 
While \siasw offers 1.7$\times$ improvement over \sindexcpu, the benefit is 56.5\% lower than that of \sindexgpu, which demonstrates the limitation of the software-only solution.
However, \siahw achieves 2.0$\times$ additional improvement over \siasw, closely approaching to \siaideal, 11.6\% higher than what \sindexgpu offers, which presents the effectiveness of hardware acceleration.
These results show the effectiveness and necessity of \sia as a solution that synergizes algorithm and hardware designs for acceleration.
% These results indicate that \sia's memoization-based algorithm in \siasw effectively reduces the computational load, while in \siahw, exploiting the FPGA accelerator alongside incremental index learning algorithm yields a high throughput closer to that of \siaideal.
%

% \begin{figure}[t]
%         \centering
%         \includegraphics[width=1.0\linewidth]{fig/eval-microbench.pdf}
%         \caption{Throughput and latency of microbenchmark with varying composition of \texttt{range}, \texttt{read}, and \texttt{insert} queries.}
%         \label{fig:microbench-throughput}
%         \vspace{-2ex}
% \end{figure}

\subsubsection{System Power Consumption}
\begin{figure}[t]
        \centering
        \includegraphics[width=0.9\linewidth]{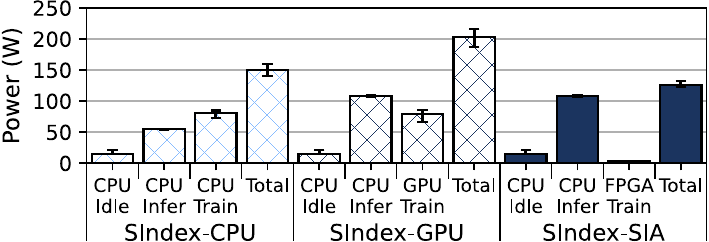}
        \vspace{-2ex}
        \caption{Average power consumption of \sindexcpu, \sindexgpu, and \sindexsia end-to-end systems. Vertical lines indicate minimum and maximum power consumption.}
        \label{fig:power_consumption}
        \vspace{-2ex}
\end{figure}

%
% While the most popular acceleration platform is GPU as of today, GPU is notorious for being very power-hungry.
% %
% Thus, GPU is not well-suited for jobs that do not have sufficient parallelism to constantly make GPU cores busy, which is the case for \lia. 
% %
% As an alternative, we choose FPGA since it allows us to customize the hardware architecture for the given task, incremental index training, offering significantly higher energy-efficiency.
%
We choose FPGA due to its capability to tailor the hardware architecture for the given task, incremental training, delivering notably higher energy efficiency compared to GPU.
Figure~\ref{fig:power_consumption} illustrates the system-level power consumption of \sindex variants: \sindexcpu, \sindexgpu, and \sindexsia, which demonstrates the advantages of FPGA acceleration in power efficiency.
We observe that the CPU-only system, \sindexcpu, operates at 150W, with a significant portion of this power attributed to CPU-based training.
%
% Note that the power consumption for inference in \sindexcpu is lower than that in other systems, as inference and training compete for CPU resources.
%
\sindexgpu operates at 203W, dissipating 79W for training at GPU and the remaining 123W for the CPU-based system.
In contrast, \sindexsia, a CPU-FPGA heterogeneous system, consumes only 126W as the FPGA accelerator adds only 3W to the CPU-only system, which demonstrates the power efficiency of the FPGA.
%
%We observe that CPU-only systems, both \sindexcpu and \liasw, operate at the same power of \highlight{150W}, since their CPUs are always busy and reach to 100\% utilization. 
%
%We also see that \sindexgpu runs at \highlight{203W}, dissipating \highlight{79W} for GPU and the remained \highlight{113W} for the CPU-based system.  
%
%Note that \sindexgpu achieves only 44\% compute utilization from GPU due to insufficient parallelism of index learning, which is why the consumed power is significantly lower than the TDP of 250W.  
%
%In contrast, \liahw, CPU-FPGA heterogeneous system, consumes only \highlight{126W} since the FPGA accelerator only uses an additional 3W on the CPU-only system. 
%

\begin{figure}[t]
        \centering
        \includegraphics[width=1.0\linewidth]{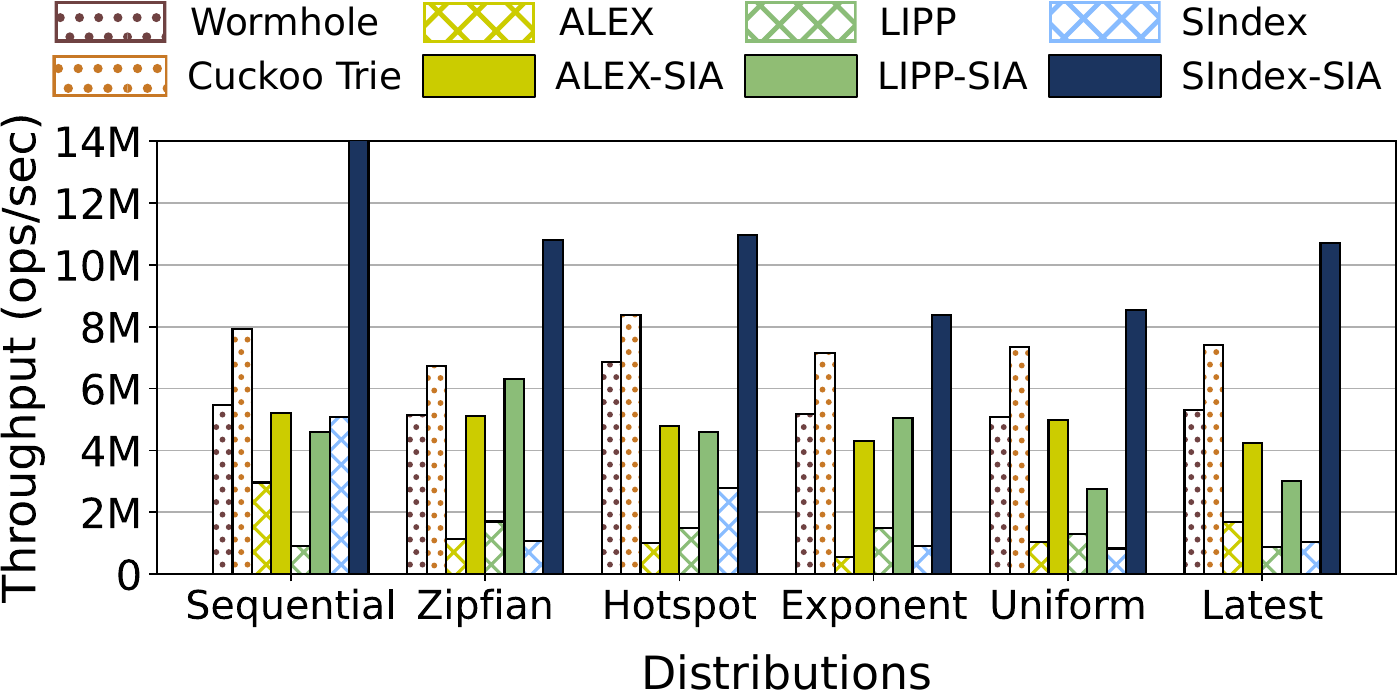}
        \vspace{-5ex}
        \caption{Throughput of \emph{non}-learned and learned indexes for queries with different request distributions.}
        \label{fig:request_dist}
        \vspace{-2ex}
\end{figure}

\subsubsection{Throughput-per-watt}
As noted in the power consumption analysis, if we only consider the accelerator itself instead of the entire system, FPGA offers 28$\times$ less power consumption compared to GPU. 
However, when we consider the system-level power consumption with their throughput together, \sindexsia achieves only 1.76$\times$ higher throughput-per-watt compared to \sindexgpu. 
While the gain may be deemed modest, the 76\% gap could translate into substantial cost disparities in terms of actual monetary expenditure since index systems tend to remain operational continuously, consistently dissipating considerable amounts of energy.
These results suggest that for the given task, the continuous retrainings of updatable learned indexes, FPGA is a more attractive option as an acceleration platform compared to GPU.

\subsubsection{Implication of Request Distributions.}
Figure~\ref{fig:request_dist} illustrates the throughput of each index across six different request distributions as used in prior works~\cite{bourbon, gadget-eurosys}: \emph{sequential}, \emph{zipfian}, \emph{hotspot}, \emph{exponent}, \emph{uniform} and \emph{latest}.
%
%
%In the case of the \emph{sequential} distribution, learned indexes demonstrate higher performance compared to other \emph{non}-learned indexes because they store keys in sorted order.
%
%Furthermore, learned indexes exhibit enhanced performance as inserted keys are clustered for skewed distributions.
%
Across all six query distributions, learned indexes accelerated with \sia consistently show significant performance improvement, which ranges from 3.9$\times$ to 6.2$\times$ in comparison with the baselines.
Note that \emph{zipfian}, \emph{hotspot}, and \emph{exponent} distributions exhibit skewed patterns, resulting in certain key ranges being accessed more frequently than others, causing more node splits.
As node splits trigger cold trainings, it imposes performance overhead, while we observe that its impact on the end throughput results is negligible.

\begin{table}[t]
        \centering
        \caption{Performance degradation as we vary training interval from 5s to 300s and deletion ratio from 5\% to 15\%.}
        \vspace{-2ex}
        \includegraphics[width=0.75\linewidth]{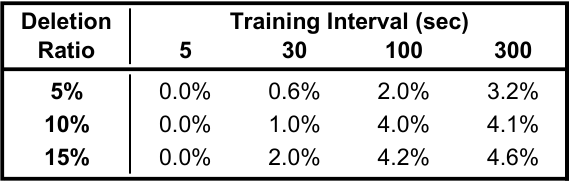}
        \label{tab:lazy-deletion-impact}
\end{table}

\begin{figure}[t]
        \centering
        \includegraphics[width=0.95\linewidth]{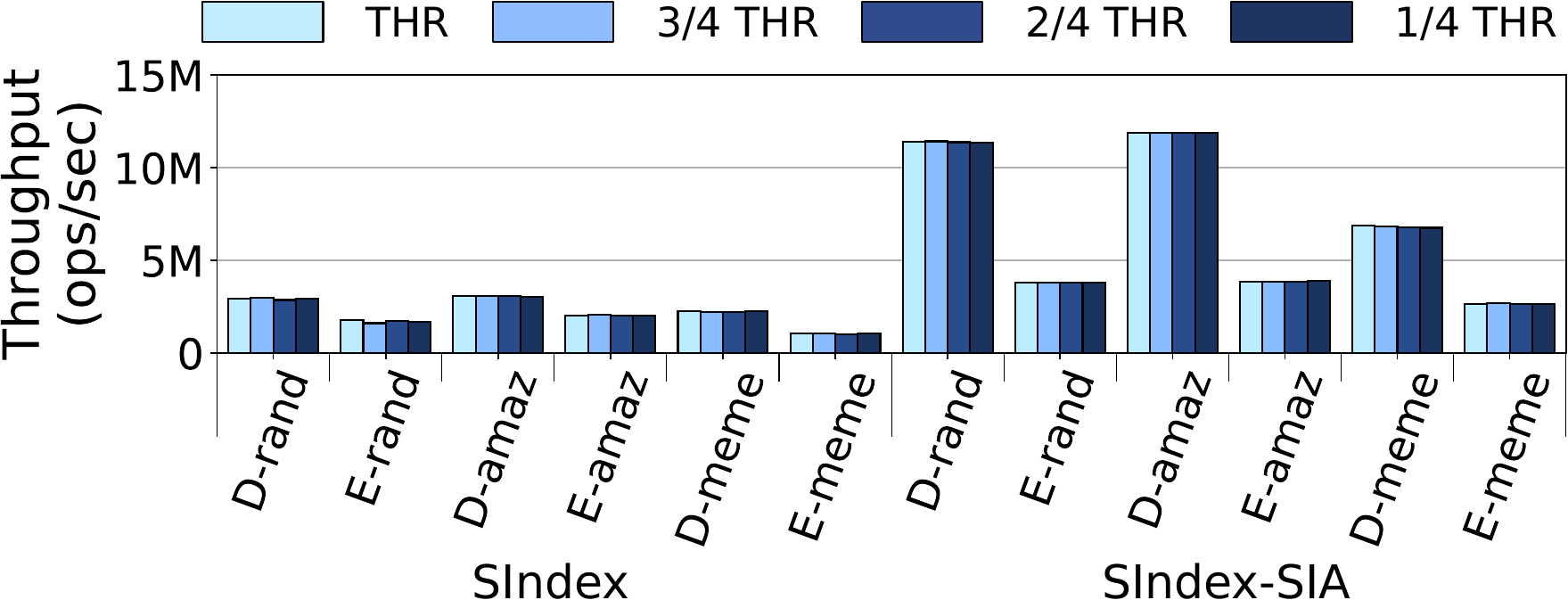}
        \vspace{-2ex}
        \caption{Throughput of \sindex and \sindexsia for YCSB workloads as the node accuracy threshold (THR) varies. As THR decreases, the learned index is more inclined to divide nodes, resulting in smaller nodes containing fewer keys.}
        \vspace{-2ex}
        \label{fig:threshold-sweep}
\end{figure}
% The evaluation begins with the original THR value, followed by assessing reduced values set at 3/4, 1/2, and 1/4 of the value.}}
%
%
\subsubsection{Implication of Lazy Delete Query Handling}
\label{sec:lazy_del_eval}
%
%
% As elaborated in Section~\ref{sec:lazy-deletion}, \lia adopts a lazy approach to eliminate deleted keys during cold training, instead of incremental learning.
%
Table~\ref{tab:lazy-deletion-impact} illustrates the impact of lazy delete query handling on the performance of \sia-accelerated learned indexes.
For this experiment, we configure the cold training interval to various durations: 5, 30, 100, and 300 seconds.
We sweep the \texttt{delete} query ratio from 5\% to 15\%, filling the remaining queries with \texttt{read} queries.
%
% Additionally, to solely evaluate the influence of lazy deletion, we set the execution time and CPU usage of cold training to be zero.
%
The results show that at a deletion ratio of 5\%, there is a performance degradation of 3.2\% when the training interval is  300 seconds.
Meanwhile, with a deletion ratio of 15\%, the larger number of unhandled keys results in a greater performance loss, which increases up to 4.6\%.
Nonetheless, the performance degradation remains at a marginal level, which validates the viability of the lazy approach, particularly considering the significant costs associated with complete cold training.

\subsubsection{Implication of Node Size on Throughput.}
\label{sec:small_leaf_nodes}
As discussed in Section~\ref{sec:split-merge}, \sia employs the same model split mechanism as in the existing system, where a node is split into two if its model accuracy drops below a certain threshold. 
To evaluate the implication, we gradually reduce the node accuracy threshold from the default value (THR) to its 1/4 (25\%) and observe the changes in throughput. 
Figure~\ref{fig:threshold-sweep} shows the results on \sindex and \sindexsia. 
As the threshold decreases, the learned indexes split the nodes more aggressively, resulting in smaller nodes with fewer keys.
While such changes incur overhead, we observe that the performance impact is negligible for both \sindex and \sindexsia, since the model splitting does not alter the total number of keys, which determines the computational load for training.  
\section{Additional Related Work}

\niparagraph{Learned index structures.}
There has been a large body of prior works~\cite{sprig, pgm-index, radixspline, lipp, plex, lisa, lidusa, multidimensional-index, tsunami, colin, histtree, bourbon, learned-index-google-deployment, li-benchmark} for learned index systems. 
%
%PGM-index~\cite{pgm-index} extends the initial learned index system by enabling predecessor and range queries within a worst-case bounds. 
% 
RadixSpline~\cite{radixspline} and PLEX~\cite{plex} further optimize learned index construction. 
%
%They eliminate multiple training passes from the original learned index by using a spline and multi-level radix layer. 
%
%While the initial learned index work focuses on 1-dimensional indexes, several works~\cite{lisa, flood, tsunami, sprig} propose to extend it and support multi-dimensional indexes. 
%
% LISA~\cite{lisa} targets spatial data by training monotonic functions that navigate to the correct shards residing in the disk.
%
Flood~\cite{multidimensional-index} and Tsunami~\cite{tsunami} exploit the learning approach for multi-dimensional indexes to automatically optimize the index structure for the given data and query distributions. 
%
%Tsunami~\cite{tsunami} builds on top of Flood to address its limitation on skewed query workloads by introducing light weight decision tree.
%
% SPRIG~\cite{sprig} proposes to build learned index with spatial interpolation function which enables the support for range and k-nearest queries.
%
% BOURBON~\cite{bourbon} builds learned index for the popular log-structured merge trees by deploying greedy piecewise linear regression models.
%
% While \lia is described for updatable string-key based learned index, it can fundamentally accelerate training for any system that employs linear regression models.
%
On the other hand, \lia optimizes training via memoized QRD algorithm enhanced by an accelerator and builds a system for integration with learned index structure.

\niparagraph{Learned index acceleration.}
%
%As learned index gained more popularity, academia recently started addressing computational challenge of learned index.
%
Colin~\cite{colin} builds and manages CPU cache-friendly learned index structure on top of PGM-index.
Colin performs key insertions in place to better utilize the caching.
Anderson et al.~\cite{anderson-aidm22} perform micro-architectural analysis of ALEX on commodity CPU and show the impact of memory hierarchy on read/write latency.
Unlike these works that aim to benefit from microarchitectural optimizations on the CPU, \lia devises iterative QRD to leverage computation reuse and further enhances the index system by offloading the training process to a separate accelerator.

\niparagraph{QR decomposition accelerator.}
As the QR decomposition makes up an essential building block of many modern applications, several architectural design for accelerators has been studied in the literature~\cite{qrdaccel-fpga18, parallel-qrd}.
Although the QRD unit is motivated from past works, none of them use these in the context of learned index systems. 
Moreover, \lia's accelerator is designed to execute multi-dimensional parallelism in the context of retraining models in learned indexes, while QRD accelerator is a small function unit.

%\niparagraph{System acceleration using hardware.}
%
%LineFS~\cite{linefs}
%
%Centaur~\cite{centaur}

%While \lia is described for updatable string-key based learned index of which most related prior works are already introduced and described in Section~\ref{sec:background}, we review the broad range of innovations in learned index technologies here. 

%Google~\cite{google-scale-database} applies learned index for their disk-based distributed database, Bigtable~\cite{bigtable} by 

%Replacement for LI: 
%HistTree~\cite{histtree}

%Benchmarking:
%Benchmarking learned indexes~\cite{li-benchmark}

\section{Conclusion}

This work offers \lia, an accelerated string-key learned index system.
These index structures require constant retraining of their machine learning models to determine the mapping between keys and their positions. 
%
%However, in data management, keys are often updated.
%
\lia mitigates the bottleneck of the current systems that incur huge overhead of training when the keys are updated.
Training observes multi-fold issues, where it is inefficient to execute on the CPU, is serial across runs as it writes to the model, and cannibalizes CPU resources from inference queries.
Based on these insights, \lia enhances the learned index training by leveraging the mathematical property that keys can be updated incrementally, and thus, can benefit from computation reuse via memoization.
\lia further boosts this training on an energy-efficient FPGA accelerator and relieves CPU resources for inference, collaboratively offering significant speedup.

\balance

\bibliographystyle{ACM-Reference-Format}
\bibliography{paper}

\end{document}